\documentclass[final]{cvpr}

\usepackage{times}
\usepackage{epsfig}
\usepackage{graphicx}
\usepackage{amsmath}
\usepackage{amssymb}
\usepackage{multirow}
\usepackage{subfigure}

\usepackage[pagebackref=true,breaklinks=true,colorlinks,bookmarks=false]{hyperref}

\def\cvprPaperID{314} 
\def\confYear{CVPR 2021}

\begin{document}

\title{Decomposition, Compression, and Synthesis (DCS)-based Video Coding: \\A Neural Exploration via Resolution-Adaptive Learning}

\author{Ming Lu\textsuperscript{\rm 1}, Tong Chen\textsuperscript{\rm 1}, Dandan Ding\textsuperscript{\rm 2}, Fengqing Zhu\textsuperscript{\rm 3}, and Zhan Ma\textsuperscript{\rm 1} \\
{\textsuperscript{\rm 1}Nanjing University, \textsuperscript{\rm 2}Hangzhou Normal University, \textsuperscript{\rm 3}Purdue University} \\
{\tt\small \{luming, tong\}@smail.nju.edu.cn, DandanDing@hznu.edu.cn, zhu0@purdue.edu, mazhan@nju.edu.cn}
}

\maketitle

\begin{abstract}
   Inspired by the facts that retinal cells actually segregate the visual scene into different attributes (e.g., spatial details, temporal motion) for respective neuronal processing, we propose to first \textbf{decompose} the input video into respective spatial texture frames (STF) at its native spatial resolution that preserve the rich spatial details, and the other temporal motion frames (TMF) at a lower spatial resolution that retain the motion smoothness; then \textbf{compress} them together using any popular video coder; and finally \textbf{synthesize} decoded STFs and TMFs for high-fidelity video reconstruction at the same resolution as its native input. This work simply applies the bicubic resampling in decomposition and HEVC compliant codec in compression, and puts the focus on the synthesis part. For resolution-adaptive synthesis, a motion compensation network (MCN) is devised on TMFs to efficiently align and aggregate temporal motion features that will be jointly processed with corresponding STFs using a non-local texture transfer network (NL-TTN) to better augment spatial details, by which the compression and resolution resampling noises can be effectively alleviated with better rate-distortion efficiency. Such ``Decomposition, Compression, Synthesis (DCS)'' based scheme is codec agnostic, currently exemplifying averaged $\approx$1 dB PSNR gain or $\approx$25\% BD-rate saving, against the HEVC anchor using reference software. In addition, experimental comparisons to the state-of-the-art methods and ablation studies are conducted to further report the efficiency and generalization of DCS algorithm, promising an encouraging direction for future video coding. 
\end{abstract}

\section{Introduction}

\begin{figure}[t]
\begin{center}
\includegraphics[width=1.0\linewidth]{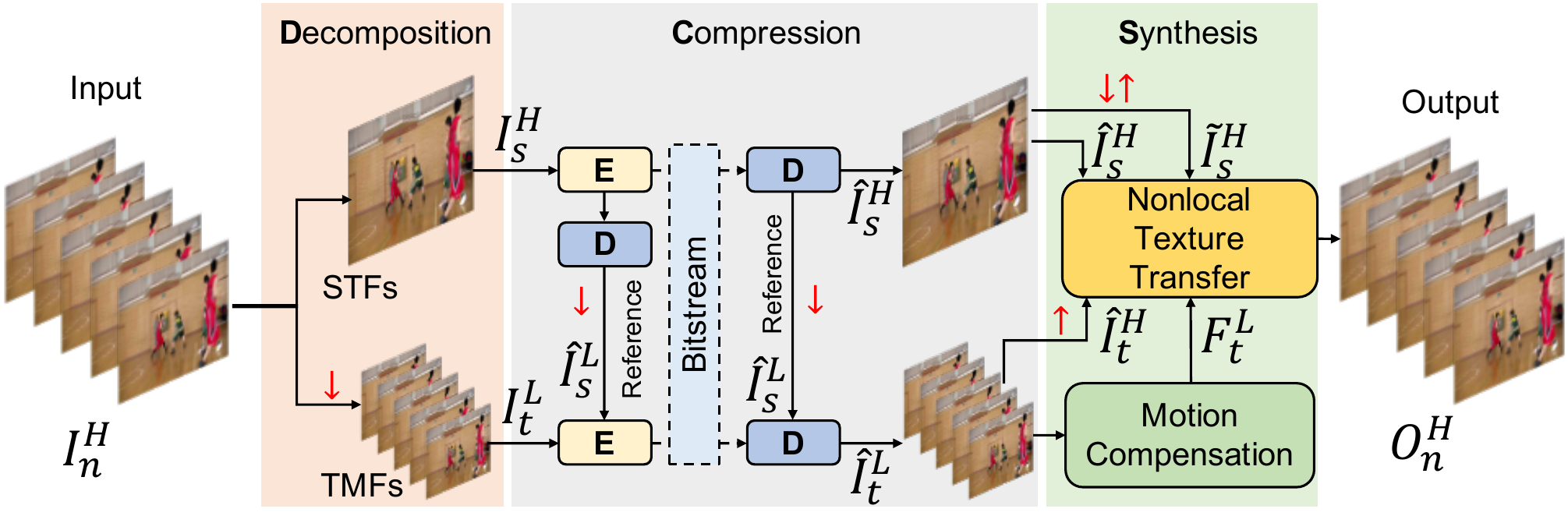}
\end{center}
   \caption{{\bf DCS-based Video Coding.} ${\downarrow}$ (in red) is for down-sampling , and $\uparrow$ (in red) is for up-sampling; {\bf E} and {\bf D} represent video encoder and decoder respectively.}
\label{fig:pipeline}
\end{figure}

Video applications have been overwhelmingly prevailing the Internet-based services, continuously striking for higher compression induced better transmission and storage. It has led to a serial well-known video coding standards in past three decades, including the H.264/Advanced Video Coding (H.264/AVC)~\cite{1218189}, H.265/High-Efficiency Video Coding (HEVC)~\cite{6316136} and emerging Versatile Video Coding (VVC). On the other hand, in past few years, built upon the advancement of deep neural networks (DNNs), a number of DNN-based video coding methods have emerged~\cite{chen2017deepcoder,lu2019dvc,rippel2019learned,liu2019learned} with encouraging compression efficiency improvements. All of these have been attempting to exploit redundancy across video frames for compact representation, where most of them assume the fixed spatiotemporal resolution granularity to process every frame. In this work, we provide an alternative solution shown in Fig.~\ref{fig:pipeline}.

\textbf{Decomposition.} The consumption of video frames over the time is the neuronal processing of natural scene stimuli in our visual pathway~\cite{schwartz2001natural}. When light photons arrive the retina, visual information will be segregated according to its attributes into different retinal cells as suggested by earlier studies. For example, cones are responsible for the photopic vision (e.g., texture details, color) while rods are more sensible for scotopic vision (e.g., illumination changes, and motion). Thus, a raw video can be decomposed as a combined representation of spatial texture and temporal motion. For simplicity, we keep using the ``frame'' term in this work, having the spatial texture frames (STFs) and temporal motion frames (TMFs) respectively.

Though we can engineer pixel-level motion displacement (e.g., optical flow) in practice, the human visual system (HVS) typically perceives the motion at semantic object basis, suggesting that a coarse block-level motion representation indeed suffices. Thus, we downscale the TMFs to a lower spatial resolution,  while keeping the STFs the same native frame size. Ideally, we wish to use STFs to preserve the spatial texture details, and have TMFs to well capture the temporal motion smoothness.

\textbf{Compression.} To support the general compatibility with existing video standards,  the STF is enforced as the intra frame of a specific group of pictures (GoP), while the rest frames in this GoP are TMFs. We choose to code STFs and low-resolution TMFs using the mainstream HEVC model in this work. Other standards, e.g., emerging VVC, can be applied as well. The STFs are HEVC intra coded, reconstructed, and downscaled to the same size as the TMF for inter prediction in Fig.~\ref{fig:pipeline}.

\textbf{Synthesis.} In a GoP, decoded STF and TMFs are used to produce high-fidelity video at its native spatiotemporal resolution. This is a {\it resolution-adaptive synthesis} problem, aiming for the high-quality restoration to alleviate compression and re-sampling noises for preserving both spatial details and temporal smoothness. Towards this goal, a motion compensation network (MCN) is first utilized to generate temporally smooth and spatially fine-grained motion representation of current TMF by aggregating the information across neighboring TMFs. This MCN is comprised of a deformable convolution network~\cite{Zhu_2019_CVPR}-based multi-frame motion feature alignment, and a separable temporal-spatial attention-based motion feature aggregation. Note that conditional convolutions are applied in this MCN to alleviate the frame quality fluctuation induced by quantization parameter (QP) bias/offset setting commonly used in video encoder, for reliable and robust model derivation. Then, temporal motion features generated by MCN, together with decoded TMFs, decoded STF as well as its re-sampled version, are fed into a non-local texture transfer network (NL-TTN) to learn and transfer cross-resolution information for high-fidelity frame restoration, where the NL-TTN has utilized high-level semantics (e.g., VGG features~\cite{simonyan2014deep}) to guide the accurate refinement for final reconstruction.

In this work, re-sampling is facilitated using the bicubic filter, and compression is compliant with the HEVC low-delay P (LDP) configuration targeting for the ultra-low-latency video applications (e.g., conferencing, tele-education). All other parameters are closely following the common test conditions suggested by the standardization committee. 
This DCS-based video coding achieves remarkable compression efficiency improvement, e.g., providing $\approx$1 dB PSNR (Peak-to-Signal-Noise Ratio) gain or $\approx$25\% BD-rate (e.g., Bjøntegaard Delta Rate~\cite{bjontegaard2001calculation}) saving against the HEVC reference model HM (\url{https://hevc.hhi.fraunhofer.de/}). On the contrary, recently emerged learnt video coding solutions~\cite{lu2019dvc,rippel2019learned,liu2019learned,Liu2020NeuralVC} only present comparable efficiency to the fast mode of HEVC compliant x265 encoder (\url{http://x265.org/}).

Extra comparisons to other state-of-the-art super-resolution-based video coding, e.g., bicubic, RCAN~\cite{Zhang_2018_ECCV}, and EDVR~\cite{Wang_2019_CVPR_Workshops},   have further revealed the superior efficiency of our DCS scheme. This is mainly because the DCS can better capture the spatiotemporal characteristics for final synthesis. Additional ablation studies have then reported the consistent robustness, and generalization of proposed solution, by carefully examining a variety of different settings, e.g., GoP size, switchable module.

Our main contributions are highlighted below:

1) We propose a novel ``decomposition, compression and synthesis'' (DCS)-based video coding framework. This work currently applies the bicubic filter-based decomposition, and HEVC compression, leaving the main focus of this study on how to efficiently devise learnt resolution-adaptive synthesis;

2) To accurately capture spatiotemporal dynamics in synthesis,  a MCN with conditional convolutions, and a NL-TTN are used for preserving both spatial details and temporal smoothness in final reconstruction.

3) This DCS-based scheme shows remarkable coding gains over the HEVC reference model, currently outperforming other learnt video coding methods. Note that our DCS generally complies with all popular video codec, promising the encouraging application prospect.

\section{Related Work}

This DCS framework is closely related to the video coding approaches utilizing the spatiotemporal re-sampling and associated super-resolution techniques.

\subsection{Re-Sampling based Video Coding}

More than a decade ago, a region-based video coding was developed in~\cite{10.1007/s11045-007-0019-y} where input high-resolution image sequences were down-sampled and compressed; Correspondingly, compressed bitstream was decoded and then super-resolved to the same resolution as its input. Note that additional region segmentation metadata classifying the motion and texture blocks, was encapsulated as the side information for more efficient decoder super-resolution. Then, key frame classification was introduced in~\cite{4711756} by which down-sampling was only enabled for non-key frames prior to being encoded. The reconstruction was fulfilled by block-matching based motion compensation and super-resolution. Similar idea was further examined in~\cite{5734821} where intra-coded frames were key frames, and inter-coded frames were non-key frames. Low-resolution inter-frames were restored using an example-based super-resolution method. More recently, a deep learning-based super-resolution was suggested by ~\cite{10.1007/978-3-030-37731-1_9} to upscale all decoded low-resolution frames for video coding. This method did not differentiate frame types and fully relied on powerful DNNs to recover re-sampling-induced missing details.

\subsection{Learnt Super-Resolution} 

The coding efficiency of re-sampling-based video coding is heavily dependent on the performance of super-resolution algorithm. In past few years, deep learning-based super-resolution methods have shown remarkable efficiency gains to conventional handcrafted algorithms. For example, a single-image super-resolution (SISR) network utilizing the residual block groups and the channel attention mechanism was developed in~\cite{Zhang_2018_ECCV}. Such SISR method was then extended in~\cite{Wang_2019_CVPR_Workshops} to use a pyramid deformable convolution network~\cite{Zhu_2019_CVPR} and an attention based network to efficiently align spatiotemporal features across frames for video super-resolution. This multi-frame super-resolution has presented clear performance lead to SISR methods by jointly exploiting the spatiotemporal correlations. Furthermore, reference-based super-resolution also attracted more attentions by which finer reference frame could be appropriately leveraged to help the super-resolution~\cite{cheng2020dual}. In~\cite{Zhang_2019_CVPR}, a texture transfer mechanism  was studied to exchange multiscale features from the reference frame for performance improvement. In the meantime, notable transformer architecture was examined in \cite{Yang_2020_CVPR} with combined hard and soft attention mechanism to extract sufficient features from the reference to improve the reconstruction.

\section{Method}

As shown in Fig.~\ref{fig:pipeline}, for every GoP, our DCS approach directly encodes the STF and low-resolution TMFs, where decoded STF is downsampled as reference to encode TMFs; Subsequently, both decoded STF and TMFs are applied to restore the video at the same resolution as the input. We mainly emphasize on the learnt resolution-adaptive synthesis in this paper, and simply utilize bicubic filter-based re-sampling and HEVC-based video compression. 

\subsection{Problem Formulation}
We use a two-frame GoP to exemplify the problem. The first frame is the STF and the second one is the TMF. The STF $I^{H}_{s}$\footnote{We use the superscripts ``H'' and ``L'' for original high resolution and downscaled low-resolution representation; and subscripts ``s'' and ``t'' for respective STF and TMF.} is first encoded with compression noise $\varepsilon^{Q}_{s}$,
\begin{equation}
\hat{I}^{H}_{s} = I^{H}_{s}  + \varepsilon^{Q}_{s}.
\label{stf_quantization}
\end{equation} Decoded STF $\hat{I}^{H}_{s}$ is then downscaled to the same resolution as the TMFs, e.g.,
\begin{equation}
\hat{I}^{L}_{s} = (\hat{I}^{H}_{s}) \downarrow_{m} + ~\varepsilon_{s}^{RS}, \label{stf_resampling} 
\end{equation} 
where $\downarrow_{m}$ represents the down-sampling operation at a factor of $m$ ($m$ = 2 in this paper for common practice), $\varepsilon^{RS}_{s}$ is the re-sampling noise. The same \eqref{stf_resampling} is applied to produce all low-resolution TMFs before compression, i.e., ${I}^{L}_{t} = ({I}^{H}_{t})\downarrow_{m} + \varepsilon_{t}^{RS}$.

Recalling that inter prediction plays a crucial role to exploit the temporal redundancy in video coding, it inherently propagates the noises accumulated in reference frame. First, the prediction of TMF from its reference STF can be simply referred to as
\begin{equation}
\tilde{I}^{L}_{t}=D(v_{t,s}){\hat I}^{L}_{s} + \varepsilon_{t,s},
\label{ref_eq}
\end{equation}
where $D(v_{t,s})$ is the 2-D matrix describing the displacement operation (e.g., warping) with  $v_{t,s}$ as the pixel-level spatial offset, and $\varepsilon_{t, s}$ is the temporal registration error due to inaccurate displacement estimation. 
Residuals between original TMF ${I}^{L}_{t}$ and its prediction $\tilde{I}^{L}_{t}$ are then compressed; and it finally leads to the reconstruction of TMF with accumulated noises:
\begin{equation}
\hat{I}^{L}_{t} = D(v_{t,s})\hat{I}^{L}_{s} + \varepsilon_{t,s} + \varepsilon^{Q}_{s} + \varepsilon^{Q}_{t}, 
\label{tmf_accumated}
\end{equation} with resampling noises $\varepsilon_{t}^{RS}$ implicitly embedded in $\hat{I}^{L}_{s}$.

In the end, we wish to restore $\hat{I}^{L}_{t}$ to its input resolution  and actively alleviate the compound noises incurred by re-sampling, quantization, as well as the inter prediction. Though we have only given an  example using a two-frame GoP illustration, it can be extended to practical scenarios with arbitrary GoP settings, where the noises will be continuously propagated over the time till the end of a GoP.

In general, the overall problem has attempted to restore decoded TMFs by transferring the high-resolution spatial details from STF and maintaining the temporal smoothness over the time in final reconstruction. Thus, we propose to solve this resolution-adaptive synthesis by multi-frame motion compensation and cross-resolution texture transfer, both of which are fulfilled using learning-based approaches. More specifically, the restored TMFs can be written as 
\begin{equation}
O^{H}_{t} = \mathcal{G}(\mathcal{F}(\{\hat{I}^{L}_{t-N}, ..., \hat{I}^{L}_{t}, ..., \hat{I}^{L}_{t+N}\}), 
\label{ref_quant_eq}
\end{equation}
with $\mathcal{F}()$ as the MCN utilizing 2$N$ neighbors for ensuring the temporal smoothness, and $\mathcal{G}()$ as the NL-TTN to learn semantic features for spatial details refinement.

\begin{figure}[t]
\begin{center}
\includegraphics[width=1.0\columnwidth]{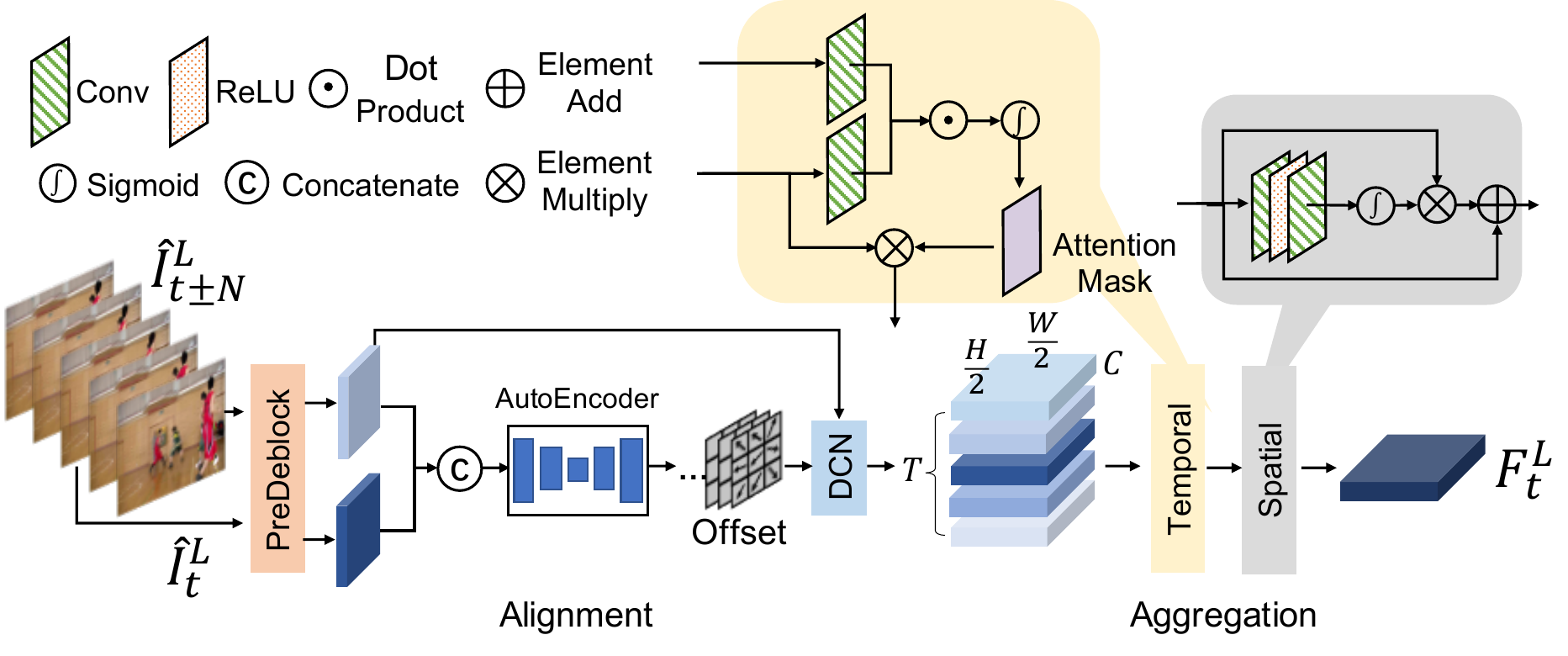}
\end{center}
   \caption{{\bf Motion Compensation Network (MCN).} A deformable convolution is used for temporal feature alignment across neighboring frames with an auto-encoder network to generate the feature offsets; and separable temporal-spatial attentions are consecutively augmented to aggregate most proper motion features. DCN stands for the deformable convolution network; T, H/2, W/2, and C are the dimensional size of feature maps. }
\label{fig:mfmc}
\end{figure}

\subsection{Learnt Resolution-Adaptive Synthesis}

\subsubsection{Framework}

We propose a learnt resolution-adaptive synthesis method to restore decoded  TMFs to their original resolution, by which the image quality is greatly enhanced at the same bit rate budget, and thus consequently improves the rate-distortion (R-D) efficiency. 

\begin{figure}[t]
\begin{center}
\subfigure[]{\includegraphics[width=0.9\columnwidth]{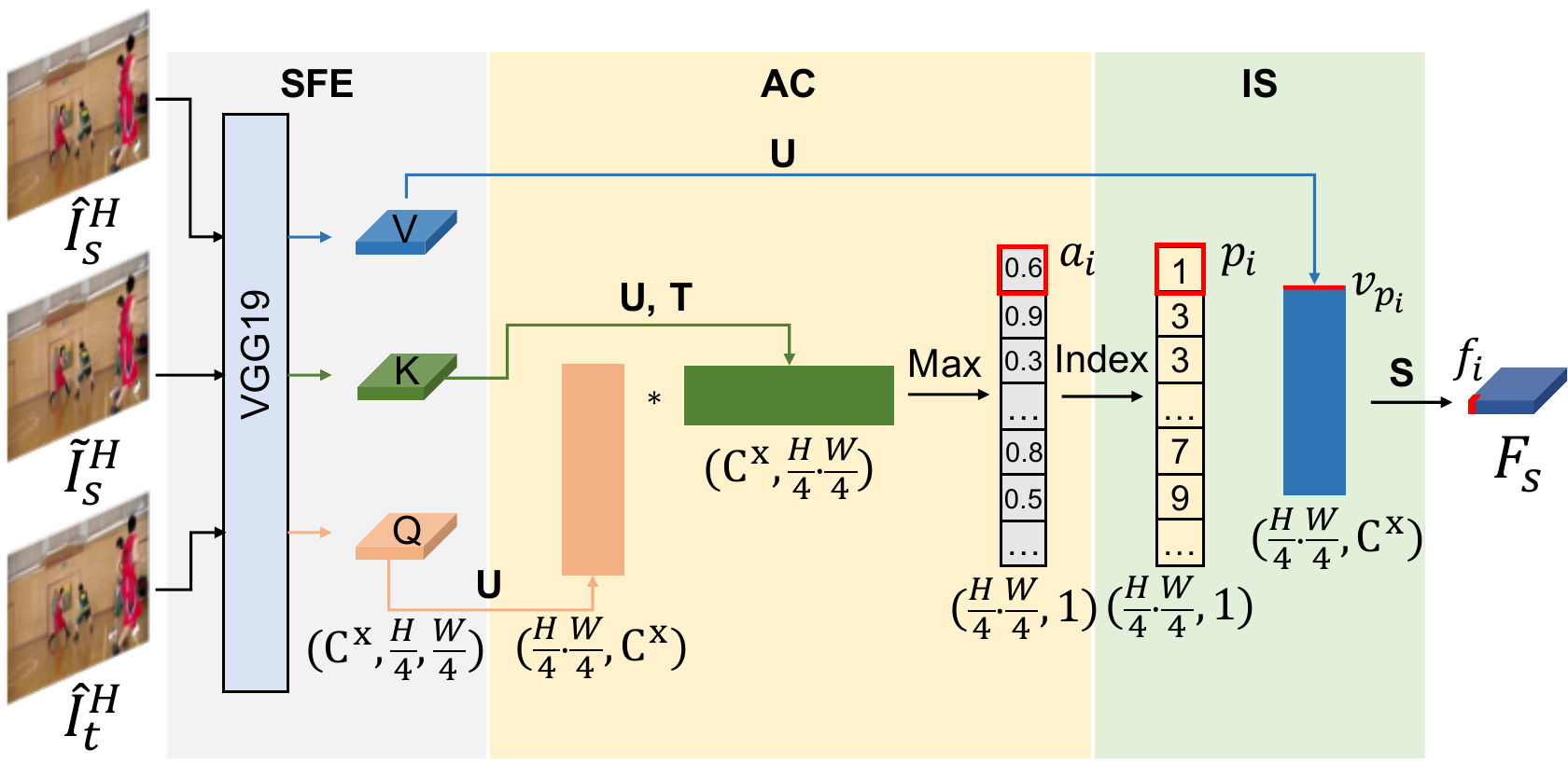} \label{ttn_a}}\\
\subfigure[]{\includegraphics[width=0.9\columnwidth]{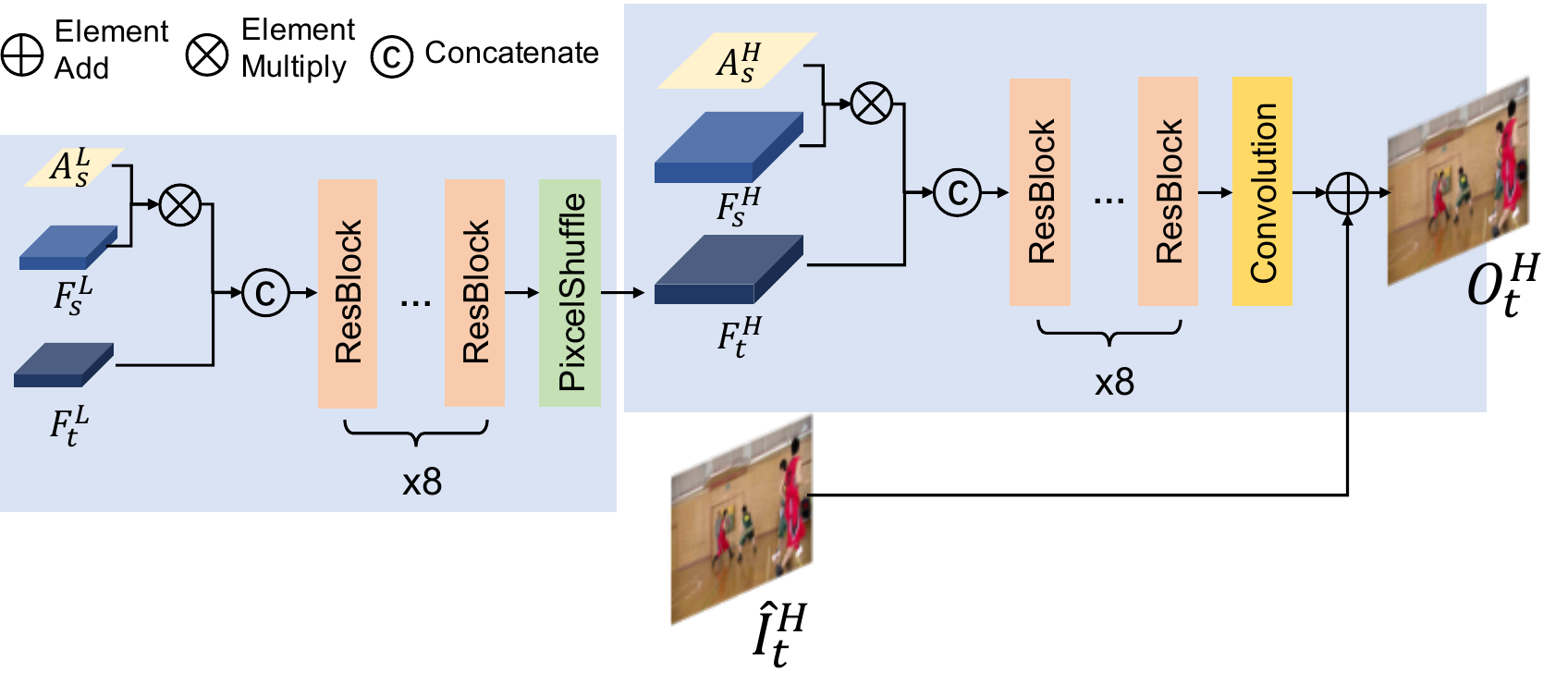}\label{ttn_b}}
\end{center}
   \caption{{\bf Non-Local Texture Transfer Network (NL-TTN).} (a) Semantically similarity generation: \textbf{T} denotes the transpose operation, \textbf{U} is for unfolding, and {\bf S} is index selection; (b) Similarity-driven fusion.}
\label{fig:ttn}
\end{figure}

To fully explore the spatiotemporal correlations to alleviate or even eliminate compound noises, we suggest a multi-frame MCN that first aligns temporal features using deformable convolutions, and then aggregates them using temporal-spatial attentions for subsequent reconstruction. Note that QP offsets are typically used across frames in practical video encoder for better R-D performance~\cite{schwarz2006analysis}. However it inevitably incurs the quality fluctuation of decoded frames when measured using the PSNR, making the model training unstable and difficult to converge. Thus, a conditional mechanism is augmented on the convolutions in MCN to tackle this issue. Furthermore, we attempt to migrate the spatial details of reference STF to TMFs for compensating the missing high-frequency components, and simultaneously retain the temporal smoothness in final reconstruction. Towards this purpose, we utilize VGG features from respective decoded STF, re-sampled STF, upscaled TMF, and aggregated motion features, in a NL-TNN for final synthesis shown in Fig.~\ref{fig:ttn}.

\subsubsection{Multi-Frame Motion Compensation}

To enforce the temporal smoothness, it is desired to accurately learn the motion dynamics across neighbor frames for compensation. It generally includes feature alignment and aggregation.

\textbf{Alignment.} In Fig.~\ref{fig:mfmc}, we set $N$ = 2, having two preceding and another two succeeding frames besides current $\hat{I}^L_t$ for a 5-frame temporal window. Recalling that multi-frame motion features are often used for temporal alignment, we choose to use the deformable convolutions~\cite{Zhu_2019_CVPR} to fulfill this purpose. A pre-deblock module consist of several convolution networks are firstly adopted to generate the deblocked features of each frame. Features of neighbor frames are concatenated with ones of the current frame respectively to input into an auto-encoder network for the prediction of the channel-wise offsets between them. With such multi-scale architecture in feature domain, we can improve the accuracy of such learnable offsets. After that, the aligned features are obtained by applying the deformable convolution layer with the neighbor features and the corresponding learned offsets as the input. Though optical flows may fit for this task, they are very vulnerable to compression noises. We align these motion features between current frame and each neighbor, initially producing $T$ stacks of feature maps at a size of $H/2 \times W/2 \times C$. $T$ is the number of neighbor frames, e.g., $T = 2N+1 = 5$; Channel dimension $C$ is 64; $H$ and $W$ are the height and width of the original input video. The details of the pre-deblock and auto-encoder networks can be found in supplementary material.

\textbf{Aggregation.} Then, we apply the attention-based mechanism to efficiently aggregate aforementioned features. We perform the separable temporal~\cite{Wang_2019_CVPR_Workshops} and spatial~\cite{Woo_2018_ECCV} attention-based aggregation sequentially. The similarity distance between aligned features and the current ones is calculated using the dot product operation on the embedded features generated from a simple convolution layer. By applying the sigmoid activation function afterwards, we can get the temporal attention mask, which can be then multiplied in a pixel-wise manner to the original aligned features. Spatial attention masks are then computed from the temporal fused features. Different from~\cite{Woo_2018_ECCV}, we remove the max-pooling layer and employ several layers depicted in the gray box of Fig.~\ref{fig:mfmc} to generate the spatial attention masks without losing the channel information. After that, the fused features are modulated by the masks through element-wise multiplication and addition.
The overall process of the MCN can be summarized as
\begin{equation}
F^{L}_{t} = \Phi(\Psi(\{\hat{I}^{L}_{t-N}, ..., \hat{I}^{L}_{t}, ..., \hat{I}^{L}_{t+N}\})),
\label{mfmc_eq}
\end{equation}
where $F^{L}_{t}$ is the aggregated motion representation features of frame $\hat{I}^{L}_{t}$ at a size of $H/2 \times W/2 \times C$, $\Psi$ represents the deformable alignment module and $\Phi$ denotes the attention-based feature aggregation.

Such deformable convolution based alignment and attention based aggregation can well capture the multi-frame spatiotemporal dynamics using pyramidal features. It could also overcome the challenging issues that could not be well handled by optical flow-based methods, such as noise, occlusion, large motion displacement, etc.~\cite{deng2020spatio}

\begin{figure}[t]
\begin{center}
\includegraphics[width=1.0\columnwidth]{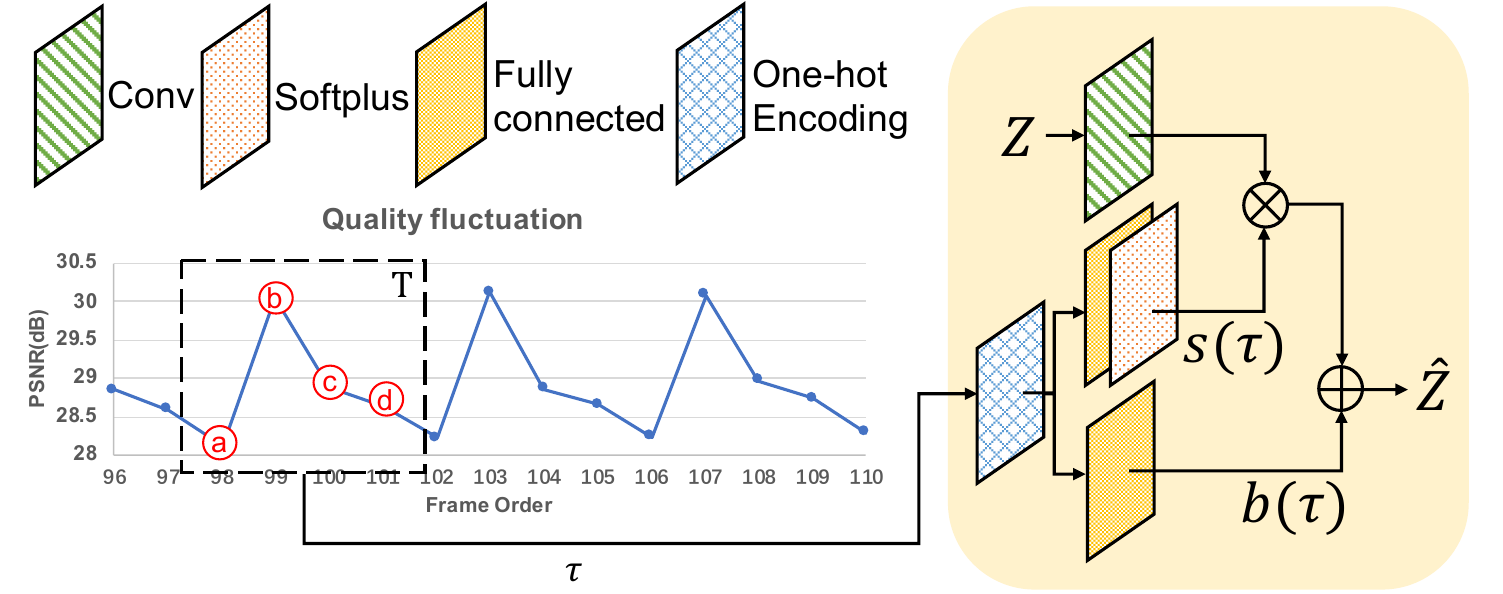} 
\end{center}
   \caption{\textbf{Conditional Convolutions.} Conditional weights and biases are trained according to the QP offset settings.}
\label{fig:condition}
\end{figure}

\textbf{Conditional Convolutions.} The QP offset is practically enforced across frames in video encoder~\cite{schwarz2006analysis} for overall R-D improvements, which is tightly connected with the prediction structures.
One example is that smaller QPs are used for reference frames, and larger QPs for other non-reference frames. It then leads to the PSNR fluctuation from one frame to another, as exemplified in Fig.~\ref{fig:condition}. Assuming a quality fluctuation pattern following the sequence of \textcircled{a}, \textcircled{b}, \textcircled{c}, and \textcircled{d}, it would incur instability if we apply the model trained when having \textcircled{b} as current TMF, to synthesize frames at \textcircled{a}, \textcircled{c} or \textcircled{d}. This is mainly because of the very different neighboring patterns induced by the QP offset. 

One solution is to train all models for all possible quality fluctuation patterns. However, it is not practical since QP offset settings may vary from one encoder configure to another. Moreover, even having all models deployed, it incurs additional overhead to switch models during synthesis from a frame to another. Inspired by the scaling factor designed for variable-rate learnt image coding~\cite{chen2020variable, Choi_2019_ICCV}, we propose to learn conditional weights and biases according to the QP offset setting that can be adaptively augmented to convolutions in both alignment and aggregation steps.

In the example shown in Fig.~\ref{fig:condition}, we attempt to learn four conditional weights and biases $\bf T$ that are defined according to this specific QP offsets. In practice, we introduce the hyper-parameter $\tau$ in training to control the conditional mechanism, e.g.,  
\begin{equation}
\hat{Z} = s(\tau)f_\theta(Z)+b(\tau), 
\label{cc_eq}
\end{equation}
where $Z$ is the input feature maps, $\hat{Z}$ is the corresponding output, $\theta$ denotes the convolution kernel and $f$ denotes the convolution operation. The channel-wise scaling factor $s(\tau)$ and additive bias term $b(\tau)$ depend on $\tau$ by
\begin{align}
s(\tau) = {\zeta}(u^{\top}{\xi}_{\mathbf{T}}(\tau)),
b(\tau) = v^{\top}{\xi}_{\mathbf{T}}(\tau),
\end{align}
where $u$ and $v$ are the fully-connected layer weights; $\top$ is the transpose operation, ${\zeta}(x) = \log(1 + e^{x})$ is the softplus function, and ${\xi}_{\mathbf{T}}(\tau)$ is onehot encoding of $\tau$ over $\mathbf{T}$~\cite{Choi_2019_ICCV}.

Noted that $s(\tau)$ and $b(\tau)$ can be retrained according to any specific QP offsets following the above methodology. We then can rewrite (\ref{mfmc_eq}) as
\begin{equation}
F^{L}_{t} = \Phi^{\tau}(\Psi^{\tau}(\{\hat{I}^{L}_{t-N}, ..., \hat{I}^{L}_{t}, ..., \hat{I}^{L}_{t+N}\})),
\label{cmfmc_eq}
\end{equation}
where {$\mathcal{F}=\Phi^{\tau}(\Psi^{\tau}())$} adapts to frame quality condition in temporal window accordingly.

\subsubsection{Non-Local Texture Transfer Network} 
To restore the spatial texture details of decoded TMFs, one may use deformable convolutions or optical flow-based warping to align STF features to current TMF for a super-resolved reconstruction. However, alignment efficiency degrades sharply when the frame distance between STF and associated TMF increases. Instead, motivated by recent reference-based super-resolution methods\cite{Zhang_2019_CVPR, Yang_2020_CVPR}, we introduce a non-local texture transfer network (NL-TTN) that resorts to semantic features for fine-grained texture transfer.

To accurately capture the texture variations induced by re-sampling and temporal displacement, we first re-sample decoded STF to mimic the re-sampling noises, and then feed $\hat{I}_{s}^{H}$ (decoded STF), $\tilde{I}_{s}^{H}$ (resampled STF) and $\hat{I}_{t}^{H}$ (upscaled TMF) altogether into a pre-trained VGG19 model~\cite{simonyan2014deep} for semantic feature extraction (SFE). It produces a set of different multiscale semantic features, a.k.a., Value (V), Key (K), and Query (Q). To reduce the computational burden, respective features at the smallest scale (e.g., at a size of $H/4\times W/4\times C^{'}$) are adopted for the following processing. $C^{'}$ equals to 256 as used in pre-trained VGG19 model.
  
We define semantic correlation between the STF and TMF using the affinity calculation (AC) between K and Q. {A $k \times k$ sliding window, e.g., $k =3$, is used to unfold K and Q features into patches at a size of [1, $C^x$] with $C^x = C'\times k\times k$, each of which can be indexed by $k_{i}, q_{j} (i, j \in [1, H/4 \times W/4])$. To fully mine the semantic information of STF, we calculate the normalized cosine similarity as affinity in a non-local manner}:
\begin{equation}
a_{i,j} = \frac{k_{i} \cdot q_{j}}{||k_{i}|| \cdot ||q_{j}||}, \label{sim_eq}
\end{equation}
where $a_{i,j}$ denotes the relevance between patch $k_{i}$ and $q_{j}$. As a result, for each $q_{j}$ in Q, the most relevant patch $k_{i}$ in K {has the relevance coefficient $a_{i} = \max_{j} a_{i,j}$ that comprises of the relevance matrix $A_s$ between Q and K.} Correspondingly, the index of $k_{i}$ is $p_{i} = \arg\max_{j} a_{i,j}$. 
We then can retrieve the most relevant patches from V, i.e., $f_{i} = v_{p_{i}}$, through index selection (IS) using the gathered indexes $p_{i}$ $(i \in [1, H/4 \times W/4])$ to construct semantically similar feature maps $F_s$. 

The $A^L_s$, $A_s^H$, $F^L_s$ and $F_s^H$ are obtained by interpolating aforementioned $A_s$ and $F_s$ at a factor of 2 to have $A^L_s$ and $F^L_s$, and at a factor of 4 to have $A^H_s$ and $F^H_s$, respectively, shown in Fig.~\ref{ttn_b}. The final restored result is obtained by fusing the motion representation $F^L_t$ with the upscaled TMF $\hat{I}_{t}^{H}$, and $F^L_s$, $F^H_s$ derived from the STF. The fusion network is consist of 8 residual blocks~\cite{He_2016_CVPR} at each scale for producing the final reconstruction
$O^{H}_{t}=\mathcal{G}(\hat{I}_{t}^{H}, F^{L}_{t}, A^{L}_{s}, A^{H}_{s}, F^{L}_{s}, F^{H}_{s})$.

\begin{table}[t]
\begin{center}
\begin{footnotesize}
\begin{tabular}{c|c|c|c}
\hline
Class & Sequence & BD-rate(\%) & BD-PSNR(dB) \\
\hline
\multirow{5}{*}{B}
    & Kimono & -29.76 & 1.12 \\
    & ParkScene & -23.33 & 0.69 \\
    & Cactus & -29.99 & 1.11 \\
    & BQTerrace & -17.54 & 0.52 \\
    & BasketballDrive & -25.90 & 0.94 \\
\hline
\multirow{4}{*}{C}
    & RaceHorses & -12.61 & 0.36 \\
    & BQMall & -22.58 & 0.93 \\
    & PartyScene & -23.48 & 0.73 \\
    & BasketballDrill & -37.28 & 1.63 \\
\hline
\multirow{4}{*}{D}
    & RaceHorses & -19.78 & 0.71 \\
    & BQSquare & -9.02 & 0.33 \\
    & BlowingBubbles & -23.74 & 0.75 \\
    & BasketballPass & -25.45 & 1.03 \\
\hline
\multirow{3}{*}{E}
    & vidyo1 & -31.57 & 1.70 \\
    & vidyo3 & -27.41 & 1.38 \\
    & vidyo4 & -30.32 & 1.53 \\
\hline
\multicolumn{2}{c|}{Average} & -24.36 & 0.97 \\
\hline
\end{tabular}
\end{footnotesize}
\end{center}
\caption{BD-rate and BD-PSNR of Proposed DCS against HEVC anchor (1-second GoP used in LDP common test)}
\label{bdbr}
\end{table}

\begin{figure}[t]
\begin{center}
\subfigure[]{
\includegraphics[width=0.45\columnwidth]{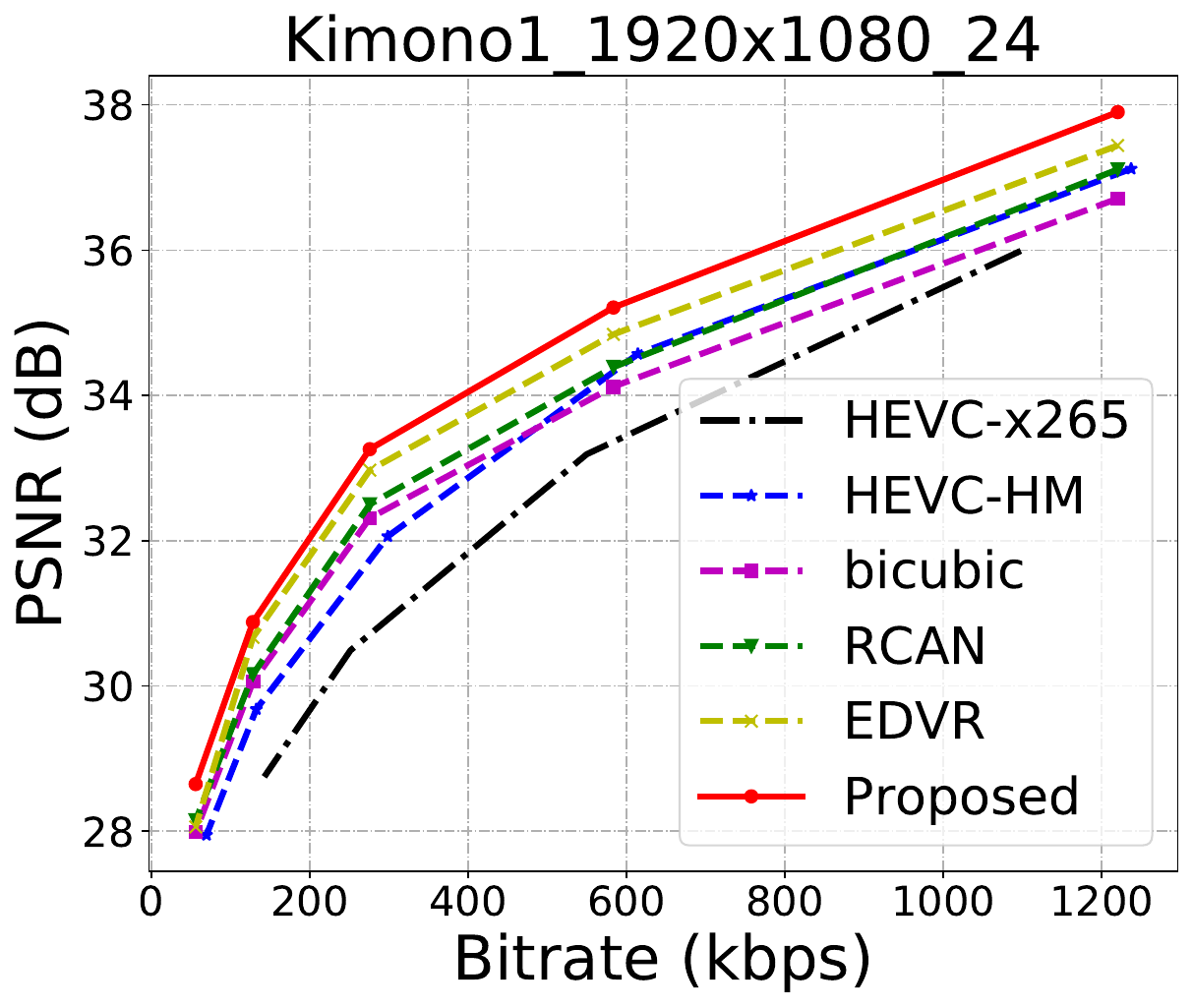}
}
\subfigure[]{
\includegraphics[width=0.45\columnwidth]{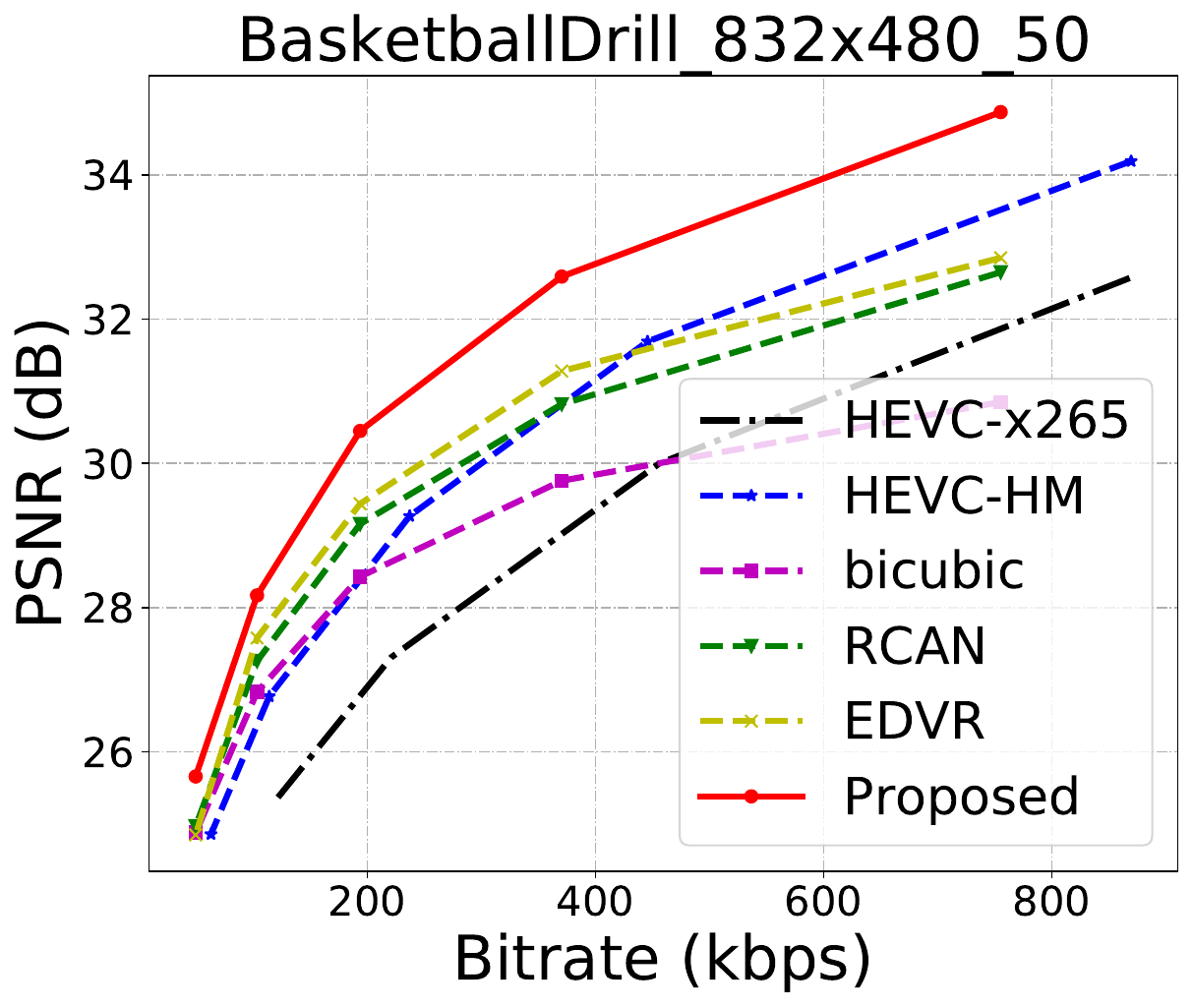}
}\\
\subfigure[]{
\includegraphics[width=0.45\columnwidth]{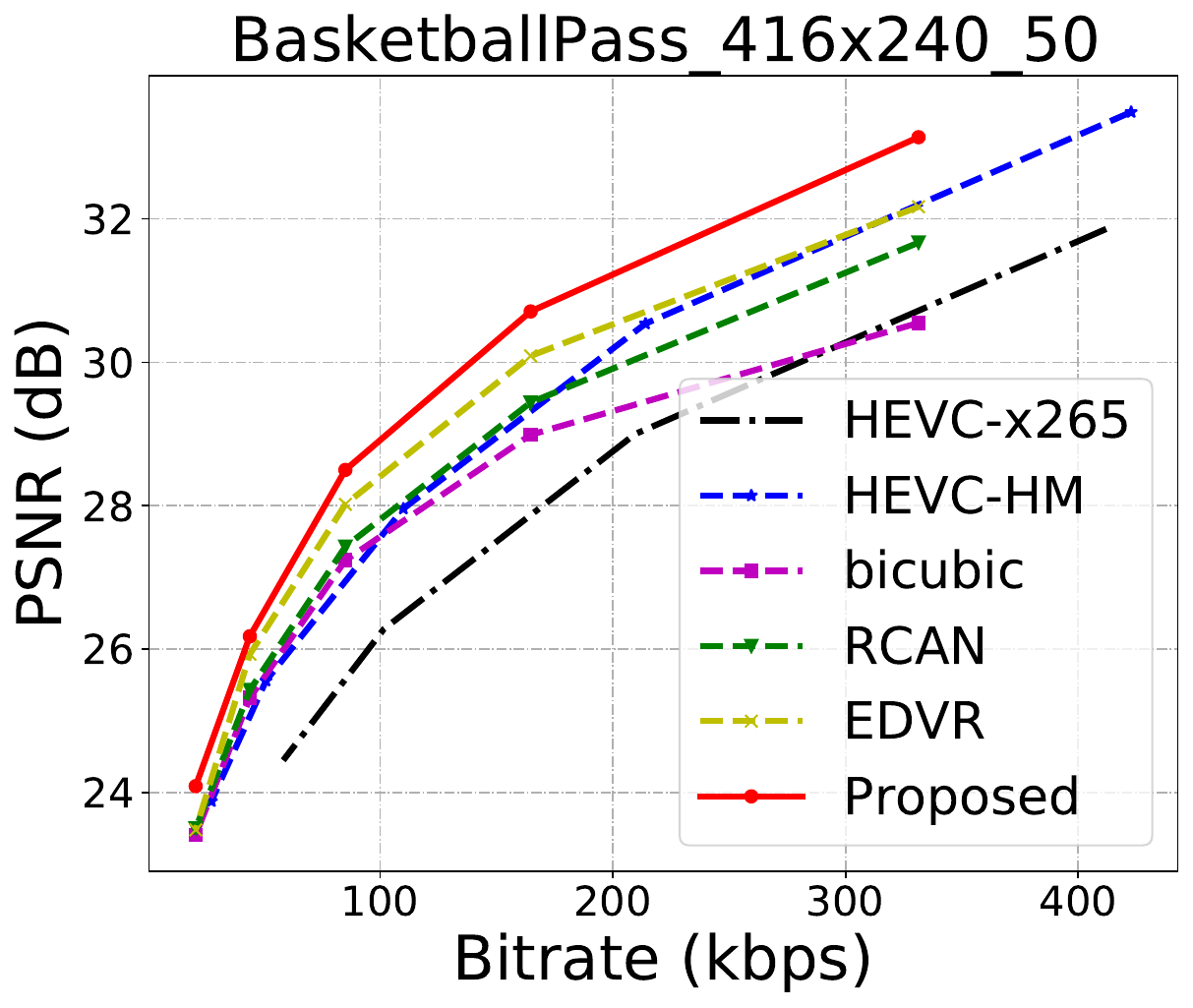}
}
\subfigure[]{
\includegraphics[width=0.45\columnwidth]{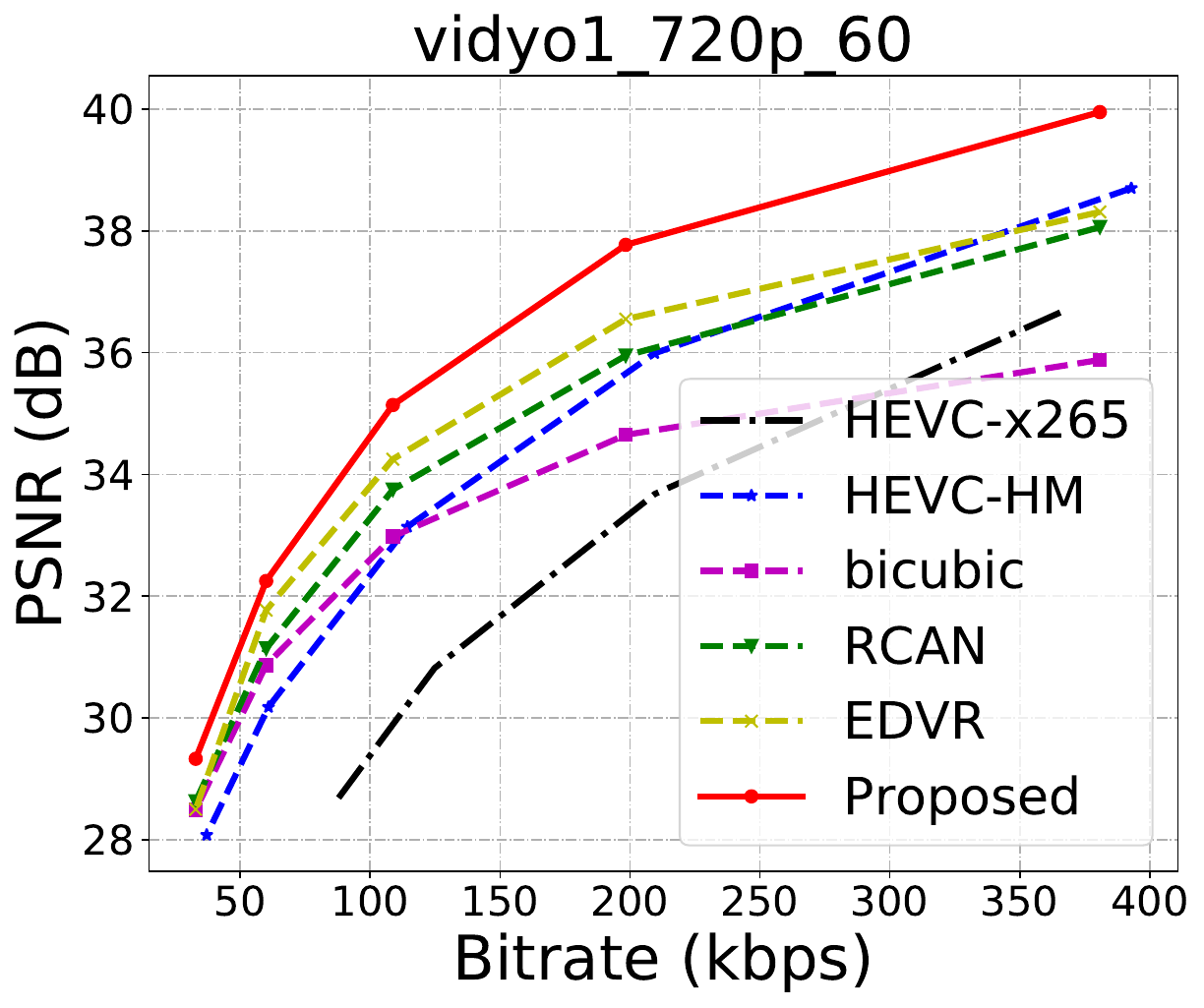}
}
\end{center}
\caption{Rate-distortion curves of proposed DCS, RCAN~\cite{Zhang_2018_ECCV}, EDVR~\cite{Wang_2019_CVPR_Workshops}, bicubic-based video coding,  HEVC anchor using reference model - HEVC-HM, and an additional HEVC encoder x265 - HEVC-x265.}
\label{compare}
\end{figure}

\section{Experimental Studies}

\subsection{Datasets and Implementation Details}

We combine the dataset provided in~\cite{8736997} that contains 182 raw videos with various resolutions and contents, and another 15 sequences released by \cite{inproceedings} for our model training. The first 100 frames of each video are used. Test evaluations are conducted using the test sequences (Class B, C, D, and E) of JCT-VC~\cite{6317156}.

As aforementioned, re-sampling operation directly utilizes the bicubic filter offered in FFmpeg\footnote{http://ffmpeg.org/}, and videos are compressed using HEVC reference software HM16.0\footnote{https://hevc.hhi.fraunhofer.de} with LDP configuration on a 2.60 GHz Intel Xeon CPU. 
Targeting for the Internet-based  ultra-low-latency video application, this work mainly emphasizes on the scenarios acrosss the medium and low bit rates. Thus, the HEVC anchors use QPs from \{32, 37, 42, 47, 51\}, while the DCS applies the QPs with matched bit rates for fair comparison. Such comparative study can be easily extended to other QPs or bitrate coverage.
The intra interval or GoP size is set to 1 second.   Other GoPs are examined in subsequent ablation studies.

In the training, the low-resolution decoded frames and original frames are randomly cropped into patches with the sizes of $64 \times 64$ and $128 \times 128$ respectively as paired training samples. Data augmentation is also done by randomly flipping in horizontal and vertical orientations, and rotating at $90^{\circ}$, $180^{\circ}$, and $270^{\circ}$. 
Adam optimizer~\cite{kingma2014adam} with default settings is used with learning rate of 1e-4. The batch size is set to 8. We conduct our experiments on PyTorch platform using Tesla V100. We train a model from scratch for each QP value, i.e., total five models are trained in this study.

Our model is trained and evaluated only in Y-channel (i.e., luminance component) of YUV space. We adopt PSNR to quantitatively evaluate the performance of DCS. The BD-rate~\cite{bjontegaard2001calculation} performance is also provided.

\subsection{Loss Function} The $L_{1}$ loss is adopted as the overall loss measurement for end-to-end training, which has been demonstrated to provide more sharp reconstruction. It is $\mathcal{L} = ||I^{H}_{t} - O^{H}_{t}||_{1}$, with $I^{H}_{t}$ for labelled ground truth, and $O^{H}_{t}$ as the spatiotemporally super-resolved results.

\subsection{Quantitative Comparison}
We first evaluate the compression efficiency of proposed DCS in comparison to the HEVC anchor using reference model HM (a.k.a., HEVC anchor for short)  . As shown in Table~\ref{bdbr}, our method consistently outperforms the standard HEVC with almost 1 dB gain (or equivalent $\approx$25\% BD-rate improvement). More than 1.5 dB gain can be observed for fast-motion ``BasketballDrive'' at 50 FPS (Frame Per Second) and slow-motion ``vidyo1'' at 60 FPS. Additionally, up to 1.12 dB improvements is also obtained for ``Kimono'' at 24 FPS. All of these have reported the generalization of DCS to videos with various content characteristics, e.g., fast/slow-motion, high/low-frame rate, rich textures, etc. 

\begin{figure*}[h]
\begin{center}
\includegraphics[width=2.0\columnwidth]{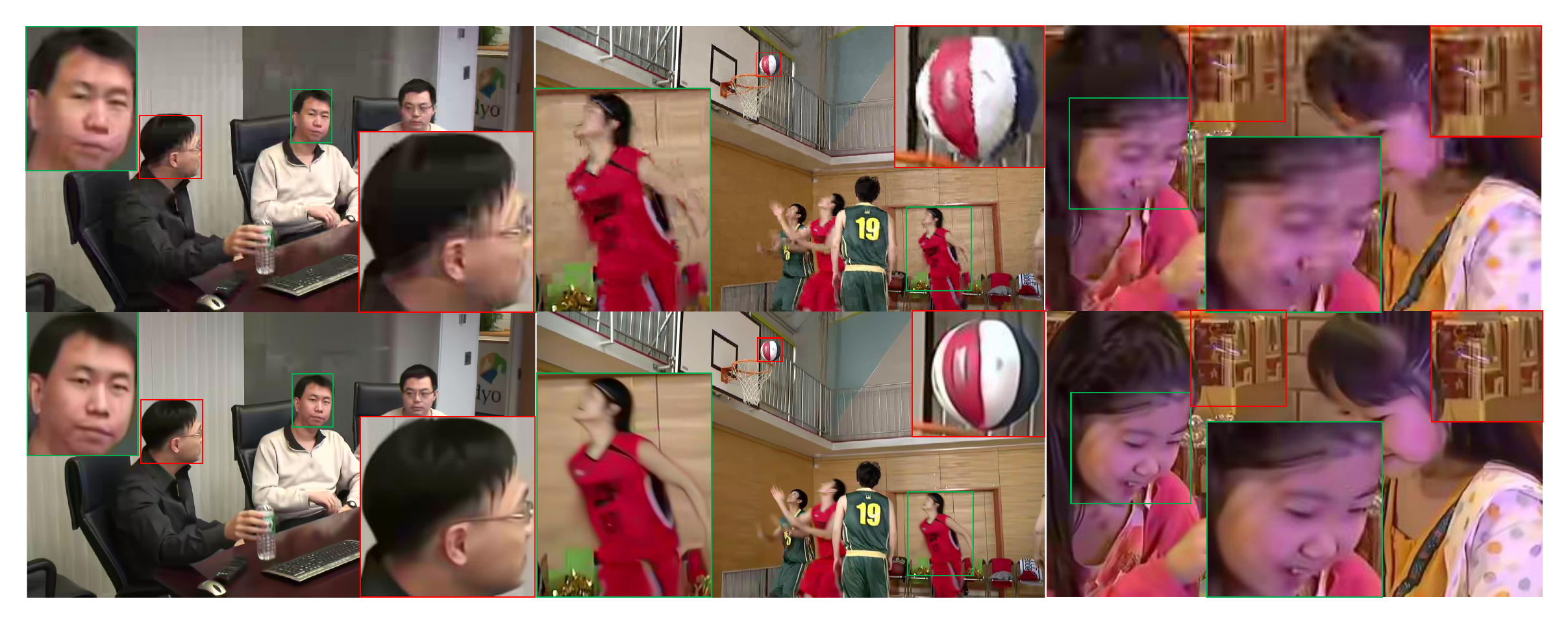}
\end{center}
   \caption{{{\bf Qualitative Visualization}: HEVC anchor (top) and proposed DCS-based method (bottom). The respective bit rate (in kbps)/PSNR (in dB) from left to right are: vidyo1 (HEVC: 115/33.14, DCS: 109/35.14), BasketballDrive (HEVC: 767/30.50, DCS: 682/31.48), BlowingBubbles (HEVC: 59/24.90, DCS: 58/26.08)}. More results can be found in supplementary materials.}
\label{fig:comparison}
\end{figure*}

We also offer comparative studies to super-resolution-based video coding using the state-of-the-art SISR method - RCAN~\cite{Zhang_2018_ECCV} and video super-resolution method - EDVR~\cite{Wang_2019_CVPR_Workshops}. In the meantime, bicubic-based upsampling is also provided as a benchmark.
Both RCAN and EDVR are re-trained using the same dataset for fair comparison. As depicted in Fig.~\ref{compare}, we select one sequence from each class for evaluation. Our DCS shows a clear performance lead  against others. Note that the lead is relatively small at critical low bit rates. In these scenarios, compression noise severely degrades the
capacity of all algorithms. However, the gains become larger when bit rate increases. It is mainly because the improved quality of STF can offer a great amount of help for the reconstruction of TMFs. Additionally, our DCS shows consistent gains across all bit rates, while existing methods only present the superior performance for a limited bit rate range, over the anchor HEVC-HM.

Additionally, x265-based HEVC encoder, a.k.a., HEVC-x265, is examined and illustrated in Fig.~\ref{compare} since most end-to-end learnt video coding methods present competitive rate-distortion efficiency to the fast mode of x265 when the distortion is measured using the PSNR metric~\cite{Liu2020NeuralVC}. It has reported about 50\% BD-Rate gain from the x265 fast mode to HM, revealing that our DCS already outperforms those end-to-end learnt approaches with a noticeable margin.

\subsection{Qualitative Visualization}
Fig.~\ref{fig:comparison} visualizes the reconstructions of both HEVC anchor and our DCS method. The selected frames have almost the same bit rates for a fair comparison with partial regions zoomed in for details. Our DCS scheme offers much better perceptual quality with sharp edges, less motion blur, and more spatial texture details.

\subsection{Complexity Discussion}
Taking the ``vidyo1'' with resolution of 720p as an example, it takes about 5514.17 seconds to encode 600 frames at  QP 42 using HEVC reference model. On the contrary, the encoding time of our DCS at corresponding bit rate is about 1157.15 seconds, having almost 80\% encoding time reduction. It is mainly because most frames in DCS are encoded as TMFs at a lower resolution. At decoder side, proposed learnt resolution-adaptive synthesis increases the processing time in addition to the HEVC decoding. Model parameters (in MBytes) and GPU running time (in seconds) are shown in Table~\ref{fig:complexity} in comparison to other super-resolution methods. Our proposed algorithm requires much less parameters, but the running time is slightly more than EDVR due to non-local affinity computation in NL-TTN. One interesting topic for next step is making current DCS more computational efficiency.

\begin{table}[t]
\begin{center}
\begin{footnotesize}
\begin{tabular}{c|c|c}
\hline
Method & param. (M) & running time (s) \\
\hline
RCAN~\cite{Zhang_2018_ECCV} & 15.44 & 0.48 \\
\hline
EDVR~\cite{Wang_2019_CVPR_Workshops} & 14.45 & 0.94 \\
\hline
Proposed & 5.63 & 1.16 \\
\hline
\end{tabular}
\end{footnotesize}
\end{center}
   \caption{Complexity comparison of DCS and other methods.}
\label{fig:complexity}
\end{table}

\section{Ablation Studies}

\begin{table}[t]
\begin{center}
\begin{footnotesize}
\begin{tabular}{c|c|c}
\hline
GOP & BD-rate(\%) & BD-PSNR(dB) \\
\hline
1s & -24.36 & 0.97 \\
\hline
2s & -28.24 & 1.12 \\
\hline
infinity & -20.63 & 0.83 \\
\hline
\end{tabular}
\end{footnotesize}
\end{center}
   \caption{Averaged BD-rate and BD-PSNR of Proposed DCS against HEVC anchor at various GoP size.}
\label{tbl:bdbr_gopsize}
\end{table}

\begin{table}[t]
\begin{center}
\begin{footnotesize}
\begin{tabular}{c|c|c}
\hline
 \multirow{2}{*}{Sequence} & BD-rate (\%) & BD-PSNR (dB) \\
\cline{2-3}
  & Ho~\cite{10.1007/978-3-030-37731-1_9} $|$ DCS & Ho~\cite{10.1007/978-3-030-37731-1_9} $|$ DCS \\
\hline
     Kimono & -14.5 $|$ -30.92 & 0.50 $|$ 1.13 \\
     ParkScene & -9.99 $|$ -27.38 & 0.24 $|$ 0.78 \\
     Cactus & -2.58 $|$ -33.40 & 0.02 $|$ 1.23 \\
     BasketballDrive & -8.09 $|$ -22.47 & 0.25 $|$  0.72 \\
\hline
\end{tabular}
\end{footnotesize}
\end{center}
   \caption{BD-rate and BD-PSNR of Proposed DCS and Ho’s Method against the HEVC anchor with infinity GoP setting.}
\label{tbl:bdbr_gop0s}
\end{table}

\textbf{GoP Size.} We further examine applications having GoP size at ``2 seconds'' or ``infinity''. For latter case, only the first frame is intra-coded STF, and all the rest frames in a sequence are inter-coded. Table~\ref{tbl:bdbr_gopsize} shows that the coding performance increases from 1-second GoP to the 2-second one, but then drops when having infinity GoP. The loss mainly comes from the scenecuts in a test sequence (e.g., camera moving in BQTerrace). It suggests the optimal coding efficiency of DCS could be possibly achieved when aligning the GoP size with scenecut. Within a scenecut, the first frame can be intra-coded STF and the rest are inter-coded TMFs. Under the same infinity GoP setting, another state-of-the-art super-resolution based video coding algorithm, e.g., Ho's method~\cite{10.1007/978-3-030-37731-1_9}, which directly down scales all frames before encoding followed by a single image super-resolution at decoder, is also included for comparison using the same HEVC anchor, shown in Table~\ref{tbl:bdbr_gop0s}. Experiments have reported the consistent performance lead ($> 2 \times$) of our DCS, even for complex-motion ``Cactus''. This further reveals the efficiency by decomposing the video into spatial and temporal attributes for separable processing and final synthesis.

\begin{figure}[t]
\begin{center}
\subfigure[]{
   \includegraphics[width=0.42\columnwidth]{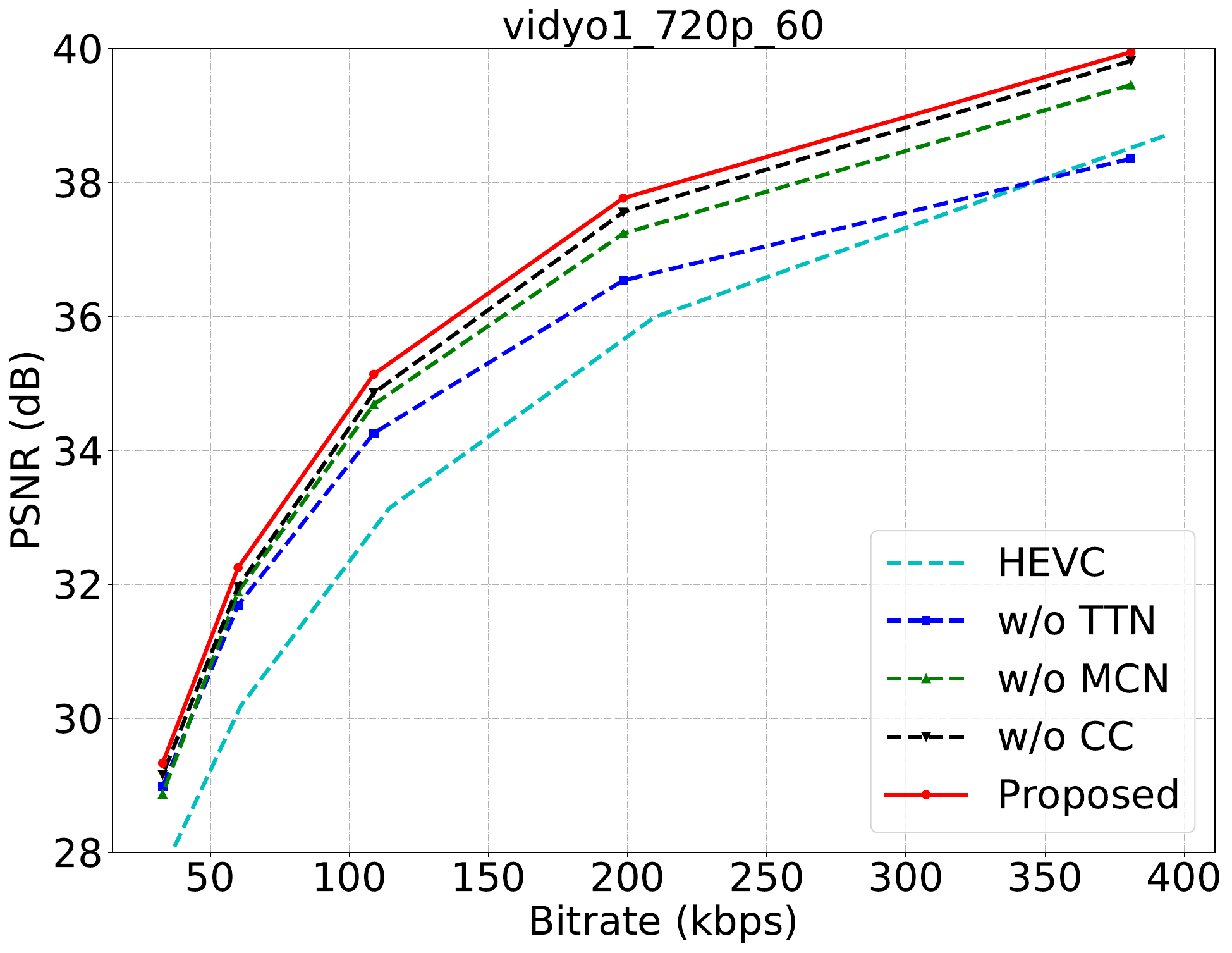}
   \label{fig:ablation_a}
   }
\subfigure[]{
   \includegraphics[width=0.5\columnwidth]{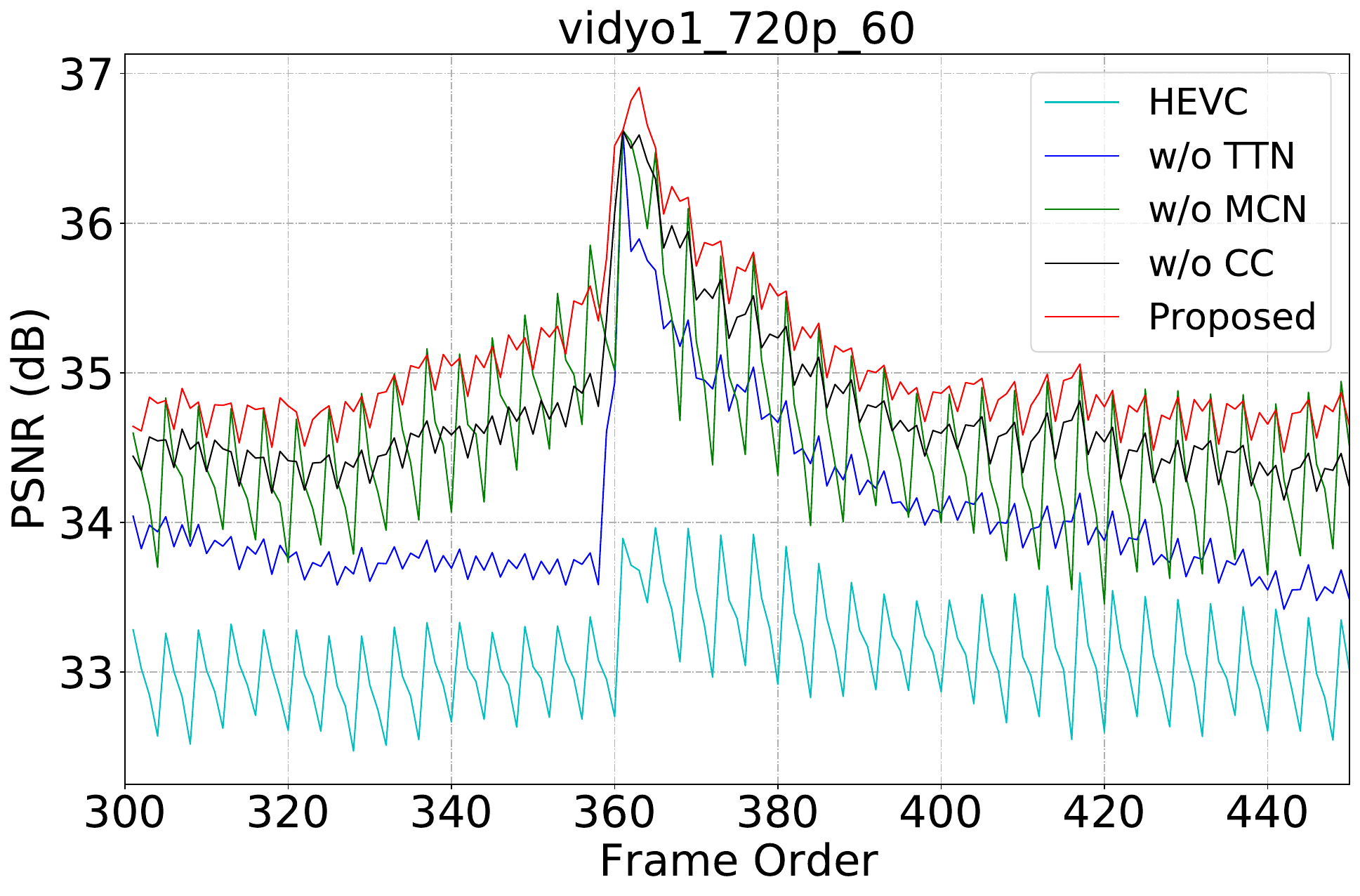} 
   \label{fig:ablation_b}
   }
\end{center}
   \caption{{\bf Module Switch.} Coding efficiency impacts of different modules in DCS. (a) averaged rate-distortion plots; (b) frame-wise snapshots; other sequences have the similar behavior.}
\label{fig:ablation}
\end{figure}

\textbf{Module Switch.} To fully understand the capabilities of our DCS, we optionally perform the module switch to study its contribution. In Fig.~\ref{fig:ablation_a}, the option ``w/o TTN'' turns off the NL-TTN, ``w/o MCN'' directly removes the MCN, ``w/o CC'' only disables the conditional convolutions in MCN, and the ``Proposed'' is the native DCS with all options enabled. In addition, we also offer the HEVC anchor for better comparative illustration. The NL-TTN offers more than 0.9 dB gains on average particularly at high bit rates; The MCN offers nearly 0.5 dB gain; Though the conditional convolutions in MCN provides $\approx$0.2 dB gain, it effectively alleviates the inter-frame quality fluctuation, making the model easier for training and faster for convergence. Frame-wise gains are consistently presented by the ``TTN'', ``MCN'', and ``CC'' in a GoP snapshot shown in Fig.~\ref{fig:ablation_b}. Our proposed model with all options enabled achieves consistent advantage across the whole bit rates for evaluation and the results show less inter frame quality fluctuation. Note that the several intermediate frames may have better PSNR by disabling the MCN, e.g., green peaks in Fig.~\ref{fig:ablation_b}. We suspect that the deformable alignment may offer the negative effect because of the temporal instability. Whereas, the overall performance is more stable when having MCN enabled. Furthermore, we have also noticed that error propagation across inter frames makes the quality drop much faster than native HEVC in Fig.~\ref{fig:ablation_b}. This may be incurred by  noises (e.g., resampling, quantization) aggregation across predictive frames, which is worth for further study.


\section{Conclusion}
The DCS algorithm  is exemplified by the bicubic filter-based decomposition, HEVC compatible video compression, and learnt resolution-adaptive synthesis. It offers an alternative and promising solution for future video coding. Currently our DCS complies with almost all popular video codec,  reporting remarkable PSNR improvements  via multi-frame motion compensation and non-local texture transfer, against the HEVC anchor and other state-of-the-art methods. Our future study will extend current framework to support more coding standards and inter prediction structures (e.g., random access, etc), investigate more advanced decomposition strategies, and explore the reduction of the computational complexity for practical application. 


{\small
\bibliographystyle{ieee_fullname.bst}
\bibliography{egbib.bib}
}

\end{document}


\title{Supplementary Materials}

\author{First Author\\
Institution1\\
Institution1 address\\
{\tt\small firstauthor@i1.org}
\and
Second Author\\
Institution2\\
First line of institution2 address\\
{\tt\small secondauthor@i2.org}
}

\maketitle


\section{Experimental Results}

\subsection{Details of BD-rate and BD-PSNR}

We detail the bitrate corresponding to each QP and the PSNR results in Table.~\ref{tb1} $\sim$ Table.~\ref{tb4} respectively for each class. Note that 1-second GoP size is used in such LDP common test. Furthermore, the DCS applies the QPs from \{27, 32, 37, 42, 47\} to generate corresponding videos with matched bit rates for fair comparison. The reason to utilize the lower QPs for the compression of our DCS method than the HEVC anchor is that the inter frames are all down-sampled within the processing of decomposition, as a result we can reduce the QP to generate the almost equivalent bit rates. The BD-rate and BD-PSNR are calculated for each sequence additionally. Lower BD-rate or higher BD-PSNR values mean a better compression for the same objective quality.

\begin{table*}[h]
\centering
\begin{scriptsize}
\caption{Comparison of Proposed DCS against HEVC anchor in Class B}
\begin{tabular}{c|c|c|c|c|c|c|c|c}
\hline
\multirow{2}{*}{Sequence} & \multicolumn{3}{c|}{HEVC} & \multicolumn{3}{c|}{DCS} & \multirow{2}{*}{BD-rate (\%)} & \multirow{2}{*}{BD-PSNR (dB)} \\
\cline{2-7}
& QP & bitrate (kb/s) & PSNR (dB) & QP & bitrate (kb/s) & PSNR (dB) & & \\
\hline
\hline
\multirow{6}{*}{Kimono}
    & 32 & 1320.75 & 37.2 & 27 & 1395.77 & 38.35 & \multirow{6}{*}{-29.76} & \multirow{6}{*}{1.12} \\
    & 37 & 661.65 & 34.66 & 32 & 678.95 & 35.9 & \\
    & 42 & 323.28 & 32.17 & 37 & 338.78 & 33.55 & \\
    & 47 & 144.09 & 29.79 & 42 & 162.25 & 31.2 & \\
    & 51 & 75.8 & 28.06 & 47 & 72.23 & 28.96 & \\
\hline
\hline
\multirow{6}{*}{ParkScene}
    & 32 & 1732.92 & 34.42 & 27 & 2017.05 & 35.53 & \multirow{6}{*}{-23.33} & \multirow{6}{*}{0.69} \\
    & 37 & 780.78 & 31.9 & 32 & 948.59 & 33.22 & \\
    & 42 & 339.11 & 29.54 & 37 & 442.79 & 31.03 & \\
    & 47 & 135.43 & 27.47 & 42 & 192.79 & 28.86 & \\
    & 51 & 65.04 & 26.02 & 47 & 77.63 & 26.92 & \\
\hline
\hline
\multirow{6}{*}{Cactus}
    & 32 & 2993.83 & 34.65 & 27 & 3073.18 & 35.41 & \multirow{6}{*}{-29.99} & \multirow{6}{*}{1.11} \\
    & 37 & 1496.15 & 32.4 & 32 & 1497.05 & 33.44 & \\
    & 42 & 758.35 & 30.04 & 37 & 751.6 & 31.31 & \\
    & 47 & 357.05 & 27.66 & 42 & 377.08 & 28.99 & \\
    & 51 & 192.71 & 25.9 & 47 & 176.29 & 26.69 & \\
\hline
\hline
\multirow{6}{*}{BQTerrace}
    & 32 & 2792.33 & 33.47 & 27 & 2948.31 & 33.5 & \multirow{6}{*}{-17.54} & \multirow{6}{*}{0.52} \\
    & 37 & 1103.95 & 31.49 & 32 & 1246.26 & 32.07 & \\
    & 42 & 487.41 & 29.22 & 37 & 586.41 & 30.41 & \\
    & 47 & 220.02 & 26.79 & 42 & 279.86 & 28.25 & \\
    & 51 & 116.7 & 24.84 & 47 & 129.14 & 26.01 & \\
\hline
\hline
\multirow{6}{*}{BasketballDrive}
    & 32 & 3347.27 & 35.45 & 27 & 3051.45 & 35.37 & \multirow{6}{*}{-25.90} & \multirow{6}{*}{0.94} \\
    & 37 & 1710.03 & 33.33 & 32 & 1475.4 & 33.58 & \\
    & 42 & 891.97 & 31.07 & 37 & 758.45 & 31.69 & \\
    & 47 & 432.95 & 28.69 & 42 & 391.54 & 29.51 & \\
    & 51 & 246.18 & 26.85 & 47 & 189 & 27.32 & \\
\hline
\end{tabular}
\label{tb1}
\end{scriptsize}
\end{table*}

\begin{table*}[h]
\centering
\begin{scriptsize}
\caption{Comparison of Proposed DCS against HEVC anchor in Class C}
\begin{tabular}{c|c|c|c|c|c|c|c|c}
\hline
\multirow{2}{*}{Sequence} & \multicolumn{3}{c|}{HEVC} & \multicolumn{3}{c|}{DCS} & \multirow{2}{*}{BD-rate (\%)} & \multirow{2}{*}{BD-PSNR (dB)} \\
\cline{2-7}
& QP & bitrate (kb/s) & PSNR (dB) & QP & bitrate (kb/s) & PSNR (dB) & & \\
\hline
\hline
\multirow{6}{*}{RaceHorses}
    & 32 & 1081.58 & 33 & 27 & 932.43 & 32.34 & \multirow{6}{*}{-12.61} & \multirow{6}{*}{0.36} \\
    & 37 & 499.05 & 30.2 & 32 & 449.47 & 30.17 & \\
    & 42 & 224.99 & 27.74 & 37 & 217 & 28.09 & \\
    & 47 & 101.26 & 25.79 & 42 & 101.46 & 26.24 & \\
    & 51 & 54.43 & 24.36 & 47 & 47.54 & 24.58 & \\
\hline
\hline
\multirow{6}{*}{BQMall}
    & 32 & 996.32 & 34.34 & 27 & 878.65 & 33.8 & \multirow{6}{*}{-22.58} & \multirow{6}{*}{0.93} \\
    & 37 & 508.31 & 31.54 & 32 & 450.07 & 31.77 & \\
    & 42 & 260.86 & 28.77 & 37 & 237.63 & 29.66 & \\
    & 47 & 125.91 & 26.13 & 42 & 123.55 & 27.17 & \\
    & 51 & 70.56 & 24.23 & 47 & 61.04 & 24.83 & \\
\hline
\hline
\multirow{6}{*}{PartyScene}
    & 32 & 1755.56 & 31.23 & 27 & 1474.06 & 30.88 & \multirow{6}{*}{-23.48} & \multirow{6}{*}{0.73} \\
    & 37 & 780.93 & 28.2 & 32 & 707.93 & 28.69 & \\
    & 42 & 329.71 & 25.39 & 37 & 339.64 & 26.46 & \\
    & 47 & 119.88 & 23 & 42 & 149.16 & 24.11 & \\
    & 51 & 50.73 & 21.5 & 47 & 56.25 & 22.18 & \\
\hline
\hline
\multirow{6}{*}{BasketballDrill}
    & 32 & 925.52 & 34.05 & 27 & 873.58 & 35.12 & \multirow{6}{*}{-37.28} & \multirow{6}{*}{1.63} \\
    & 37 & 477.45 & 31.59 & 32 & 433.77 & 32,81 & \\
    & 42 & 253.93 & 29.18 & 37 & 228.83 & 30.64 & \\
    & 47 & 123.09 & 26.68 & 42 & 122.62 & 28.31 & \\
    & 51 & 67.14 & 24.71 & 47 & 58.93 & 25.81 & \\
\hline 
\end{tabular}
\label{tb2}
\end{scriptsize}
\end{table*}

\begin{table*}[h]
\centering
\begin{scriptsize}
\caption{Comparison of Proposed DCS against HEVC anchor in Class D}
\begin{tabular}{c|c|c|c|c|c|c|c|c}
\hline
\multirow{2}{*}{Sequence} & \multicolumn{3}{c|}{HEVC} & \multicolumn{3}{c|}{DCS} & \multirow{2}{*}{BD-rate (\%)} & \multirow{2}{*}{BD-PSNR (dB)} \\
\cline{2-7}
& QP & bitrate (kb/s) & PSNR (dB) & QP & bitrate (kb/s) & PSNR (dB) & & \\
\hline
\hline
\multirow{6}{*}{RaceHorses}
    & 32 & 317.35 & 32.31 & 27 & 294.68 & 32.47 & \multirow{6}{*}{-19.78} & \multirow{6}{*}{0.71} \\
    & 37 & 153.93 & 29.42 & 32 & 143.49 & 29.93 & \\
    & 42 & 77.63 & 27.1 & 37 & 71.13 & 27.66 & \\
    & 47 & 36.44 & 24.94 & 42 & 36.84 & 25.61 & \\
    & 51 & 20.03 & 23.33 & 47 & 18.2 & 23.64 & \\
\hline
\hline
\multirow{6}{*}{BQSquare}
    & 32 & 388.81 & 31.32 & 27 & 336.06 & 30.43 & \multirow{6}{*}{-9.02} & \multirow{6}{*}{0.33} \\
    & 37 & 167.21 & 28.49 & 32 & 160.19 & 28.58 & \\
    & 42 & 77.33 & 25.62 & 37 & 82.03 & 26.58 & \\
    & 47 & 36.38 & 22.83 & 42 & 43.11 & 23.8 & \\
    & 51 & 20.2 & 20.79 & 47 & 21.86 & 21.23 & \\
\hline
\hline
\multirow{6}{*}{BlowingBubbles}
    & 32 & 410.32 & 31.06 & 27 & 343.25 & 31.09 & \multirow{6}{*}{-23.74} & \multirow{6}{*}{0.75} \\
    & 37 & 180.65 & 28.11 & 32 & 167.19 & 28.79 & \\
    & 42 & 76.77 & 25.53 & 37 & 79.86 & 26.59 & \\
    & 47 & 29.79 & 23.33 & 42 & 36.87 & 24.33 & \\
    & 51 & 14.88 & 22.01 & 47 & 16.28 & 22.52 & \\
\hline
\hline
\multirow{6}{*}{BasketballPass}
    & 32 & 433.84 & 33.54 & 32 & 364.1 & 33.44 & \multirow{6}{*}{-25.45} & \multirow{6}{*}{1.03} \\
    & 37 & 220.79 & 30.59 & 32 & 183.34 & 30.95 & \\
    & 42 & 113.35 & 28.01 & 37 & 95.41 & 28.65 & \\
    & 47 & 52.9 & 25.62 & 42 & 49.83 & 26.38 & \\
    & 51 & 23.94 & 23.94 & 47 & 23.98 & 24.23 & \\
\hline  
\end{tabular}
\label{tb3}
\end{scriptsize}
\end{table*}

\begin{table*}[t]
\centering
\begin{scriptsize}
\caption{Comparison of Proposed DCS against HEVC anchor in Class E}
\begin{tabular}{c|c|c|c|c|c|c|c|c}
\hline
\multirow{2}{*}{Sequence} & \multicolumn{3}{c|}{HEVC} & \multicolumn{3}{c|}{DCS} & \multirow{2}{*}{BD-rate (\%)} & \multirow{2}{*}{BD-PSNR (dB)} \\
\cline{2-7}
& QP & bitrate (kb/s) & PSNR (dB) & QP & bitrate (kb/s) & PSNR (dB) & & \\
\hline
\hline
\multirow{6}{*}{vidyo1}
    & 32 & 446.98 & 38.8 & 27 & 470.8 & 40.2 & \multirow{6}{*}{-31.57} & \multirow{6}{*}{1.70} \\
    & 37 & 242.02 & 36.11 & 32 & 253.37 & 38.07 & \\
    & 42 & 132.9 & 33.28 & 37 & 142.06 & 35.48 & \\
    & 47 & 70.97 & 30.33 & 42 & 79.31 & 32.59 & \\
    & 51 & 43.17 & 28.22 & 47 & 43.16 & 29.63 & \\
\hline
\hline
\multirow{6}{*}{vidyo3}
    & 32 & 522.47 & 38.43 & 27 & 541.04 & 39.33 & \multirow{6}{*}{-27.41} & \multirow{6}{*}{1.38} \\
    & 37 & 279.63 & 35.67 & 32 & 285.77 & 37.07 & \\
    & 42 & 148.96 & 32.62 & 37 & 157.8 & 34.45 & \\
    & 47 & 73.46 & 29.4 & 42 & 83.15 & 31.48 & \\
    & 51 & 39.25 & 27.22 & 47 & 40.66 & 28.48 & \\
\hline
\hline
\multirow{6}{*}{vidyo4}
    & 32 & 423.54 & 38.39 & 27 & 458.8 & 39.81 & \multirow{6}{*}{-30.32} & \multirow{6}{*}{1.53} \\
    & 37 & 223.11 & 35.84 & 32 & 243.02 & 37.68 & \\
    & 42 & 122.9 & 33.16 & 37 & 134.67 & 35.28 & \\
    & 47 & 66.38 & 30.46 & 42 & 75.85 & 32.58 & \\
    & 51 & 40.53 & 28.41 & 47 & 40.87 & 29.85 & \\
\hline
\end{tabular}
\label{tb4}
\end{scriptsize}
\end{table*}
\FloatBarrier

\subsection{Qualitative Visualization: More Results}

We give more qualitative visualization results with partial regions zoomed in for details of both HEVC anchor (left) and our DCS method (right) in this section. The respective bitrate (in kbps) and PSNR (in dB) results are attached under the corresponding pictures. 

\begin{figure*}[h]
\centering
\includegraphics[width=1.0\columnwidth]{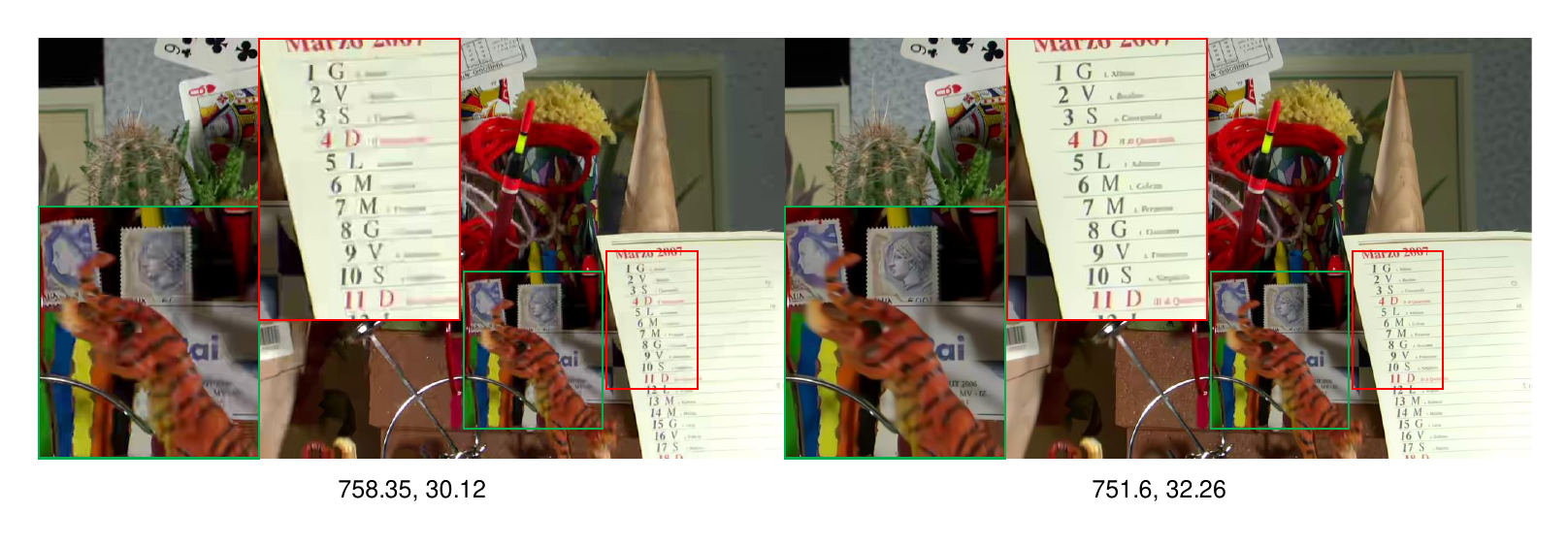} 
\caption{Catus (1920$\times$1080@50Hz)}
\end{figure*}

\begin{figure*}[h]
\centering
\includegraphics[width=1.0\columnwidth]{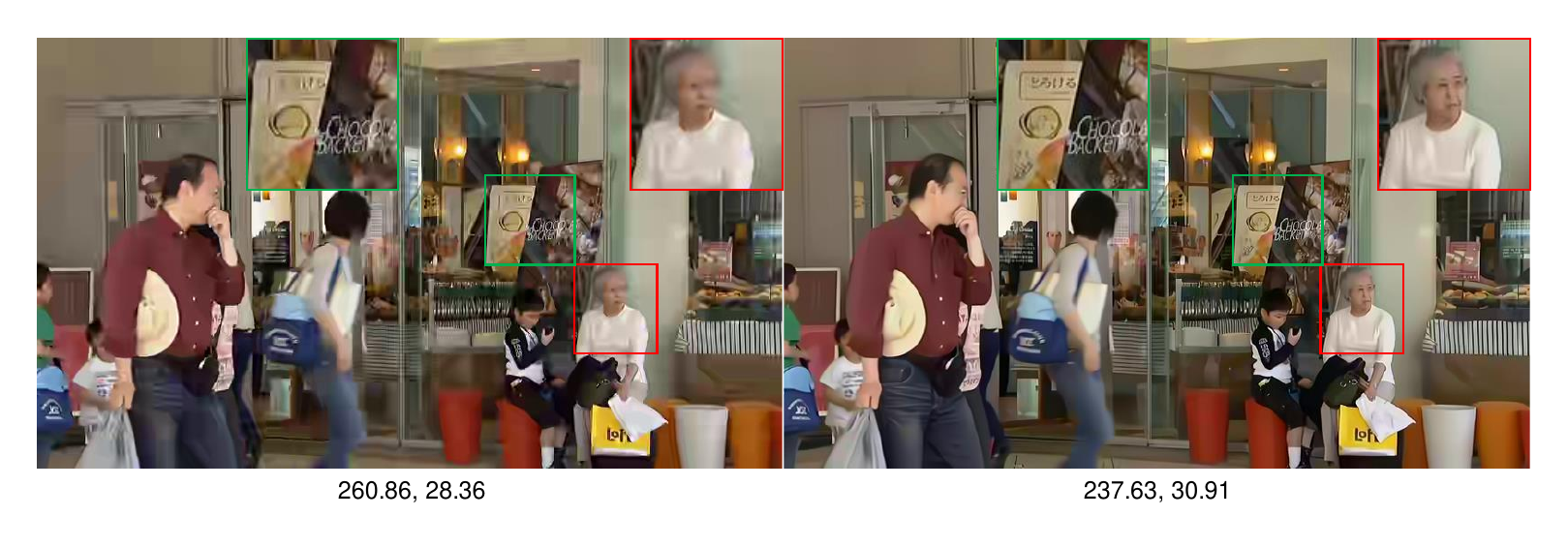} 
\caption{BQMall (832$\times$480@60Hz)}
\end{figure*}

\begin{figure*}[h]
\centering
\includegraphics[width=1.0\columnwidth]{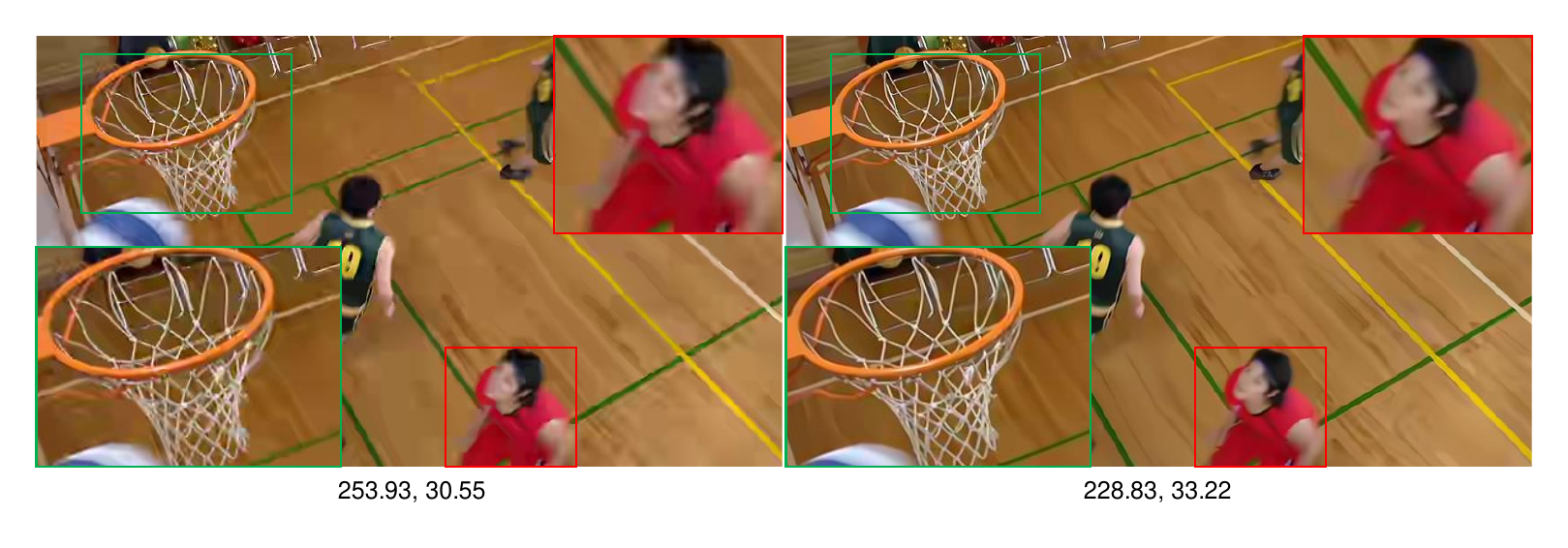} 
\caption{BasketballDrill (832$\times$480@50Hz)}
\end{figure*}

\begin{figure*}[h]
\centering
\includegraphics[width=1.0\columnwidth]{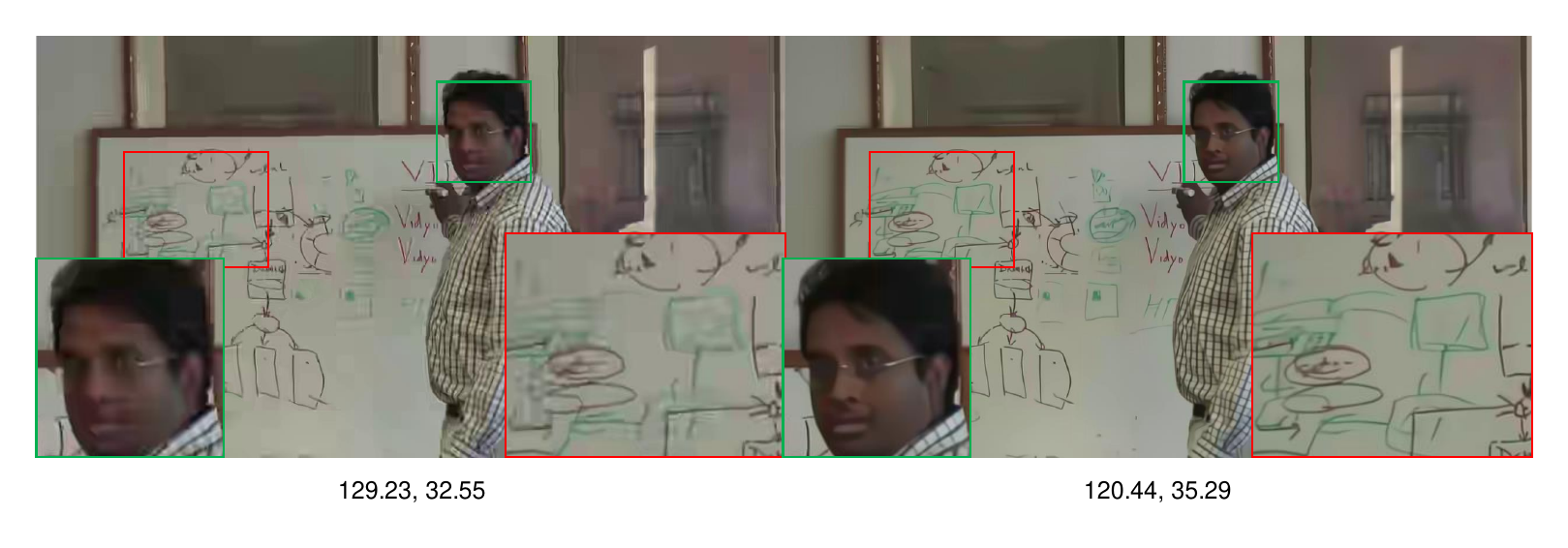} 
\caption{Vidyo3 (1280$\times$720@60Hz)}
\end{figure*}

\begin{figure*}[h]
\centering
\includegraphics[width=1.0\columnwidth]{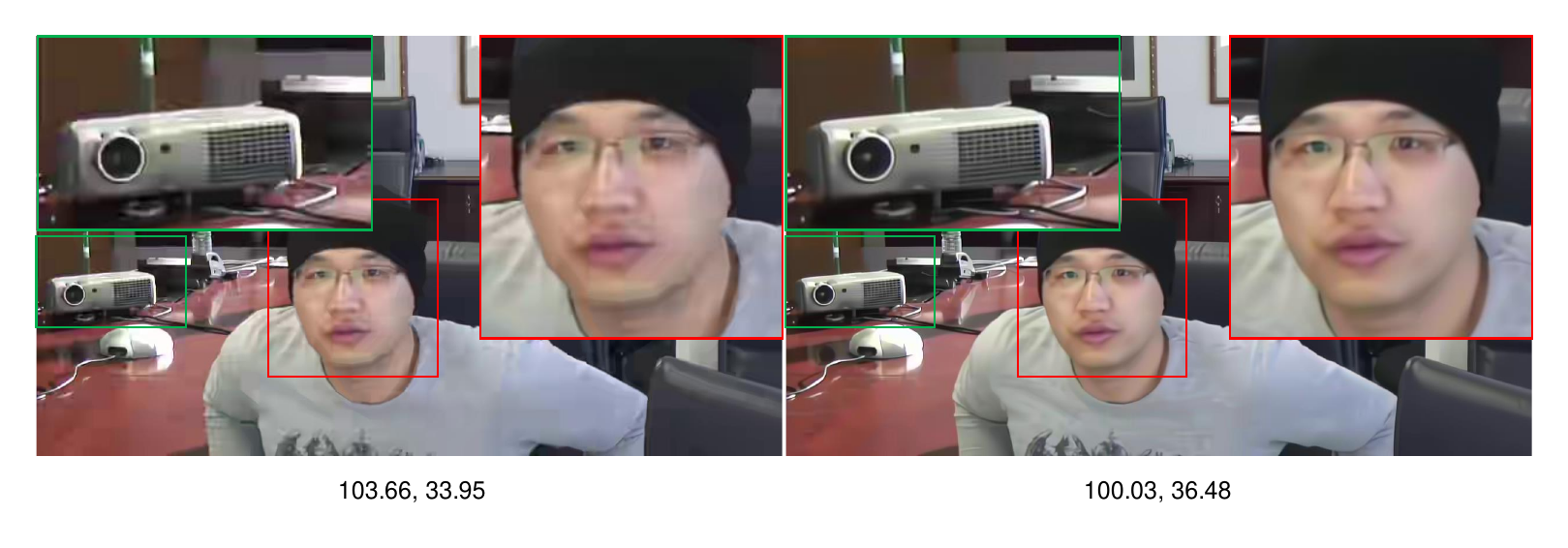} 
\caption{Vidyo4 (1280$\times$720@60Hz)}
\end{figure*}
\FloatBarrier

\section{Network Architecture}

\begin{figure*}[h]
\centering
\includegraphics[width=0.7\columnwidth]{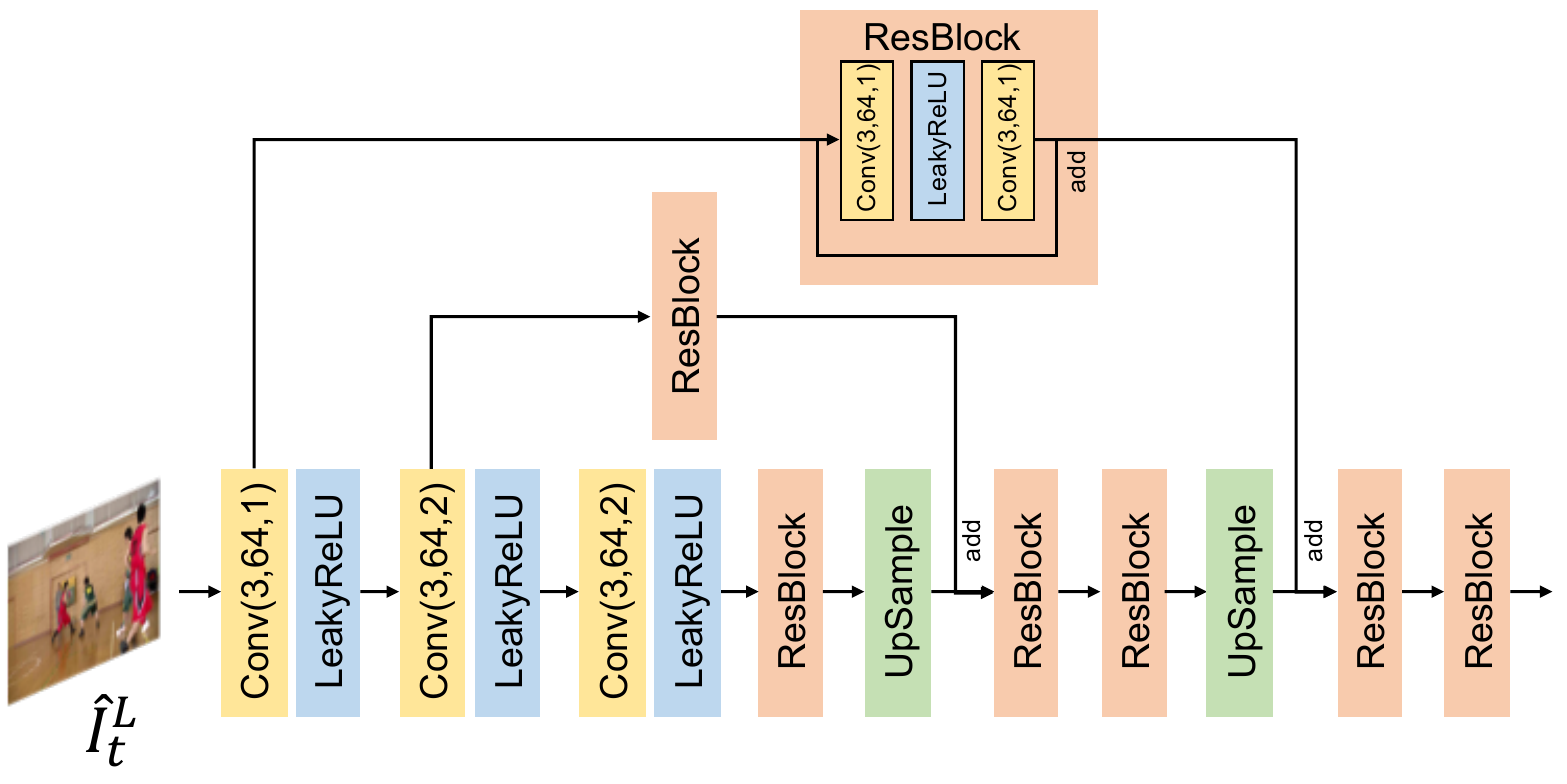} 
\caption{Network architecture of the pre-deblock network. The convolutional layers are denoted in the form of Conv(kernel size, output channels, stride size). The interpolation operation, i.e., bilinear is used as the upsampling layer.}
\label{fig:predeblock}
\end{figure*}

\begin{figure*}[h]
\centering
\includegraphics[width=0.8\columnwidth]{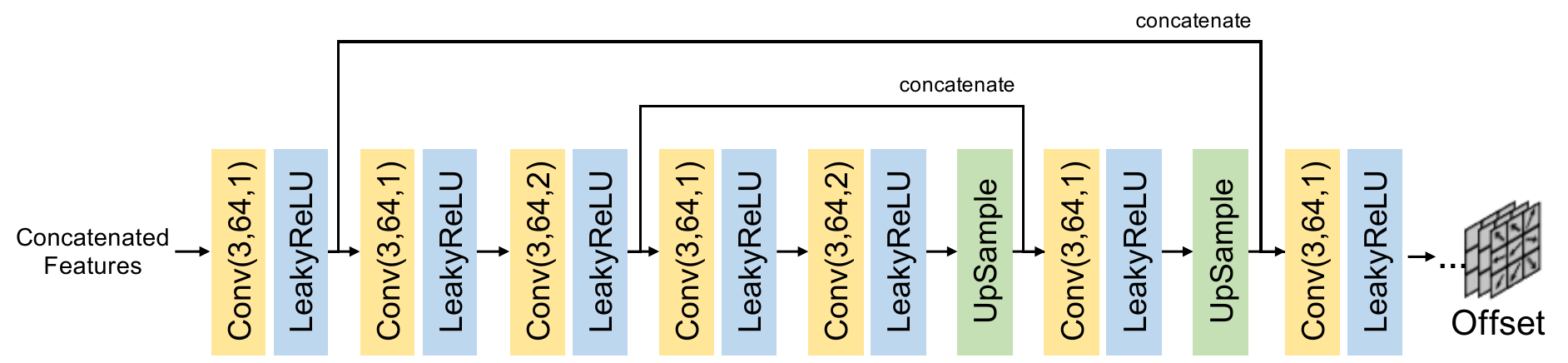} 
\caption{Network architecture of the auto-encoder network. The pre-deblocked features of the current frame and its neighboring frame are concatenated to input into the network.}
\label{fig:autoenc}
\end{figure*}

The pre-deblock module of the MCN is shown in Fig.~\ref{fig:predeblock}. And the auto-encoder network to generate the offsets of the MCN is shown in Fig.~\ref{fig:autoenc}. More details about the networks (i.e., MCN, NL-TTN) of the LRAS can be seen in attached codes. 

